\documentclass[]{fairmeta} 

\usepackage{amsmath,amsfonts,bm}

\def\Secref#1{Section~\ref{#1}}

\def\eqref#1{equation~\ref{#1}}

\def\ceil#1{\lceil #1 \rceil}

\def\1{\bm{1}}

\DeclareMathAlphabet{\mathsfit}{\encodingdefault}{\sfdefault}{m}{sl}
\SetMathAlphabet{\mathsfit}{bold}{\encodingdefault}{\sfdefault}{bx}{n}

\newcommand{\softmax}{\mathrm{softmax}}

\newcommand{\pth}[1]{\left( {#1} \right)}

\newcommand{\nmax}{N_{\mathrm{max}}}

\newcommand{\iha}{\textbf{IHA}}
\newcommand{\mha}{\textbf{MHA}}

\usepackage{hyperref}
\usepackage{mathtools}
\usepackage[inline]{enumitem}
\usepackage{graphicx}
\usepackage{url}
\usepackage{subfig}
\usepackage{nicefrac}
\usepackage{cleveref}
\usepackage[table]{xcolor} 
\usepackage{array}         
\usepackage{booktabs}  
\usepackage{bbm}

\usepackage{algorithm}
\usepackage{algpseudocode} %

\usepackage{amsfonts}       
\usepackage{nicefrac}       
\usepackage{microtype}      
\usepackage{xcolor}         
\usepackage{amsmath}
\usepackage{amsfonts}
\usepackage{amssymb}
\usepackage{amsthm}
\usepackage{enumitem}
\usepackage{xcolor}
\usepackage{graphicx}
\usepackage{algorithm}
\usepackage{algpseudocode}
\usepackage[most]{tcolorbox}
\usepackage{natbib}
\usepackage{subcaption} 

\usepackage{graphicx}
\usepackage{subcaption}
 \usepackage{float}

\crefname{equation}{Eq.}{Eqs.}
\Crefname{equation}{Eq.}{Eqs.}

\definecolor{nblue}{HTML}{D7E5FF}

\definecolor{ngreen}{HTML}{E7FCF2}

\definecolor{nred}{HTML}{FFE8ED}

\newtcolorbox[auto counter]{theorembox}[2][]{%
  colframe=nblue,
  colback=nblue,
  breakable,
  before upper={{\textbf{Theorem \thetcbcounter.} #2}},
  #1
}
\makeatletter
\def\tcb@cnt@theoremboxautorefname{Thm.}
\makeatother

\newtcolorbox[auto counter]{mainbox}[2][]{%
  colframe=ngreen, 
  colback=ngreen,  
  before upper={{\textbf{Main Contributions.} #2}},
  #1
}

\newtcolorbox[auto counter]{qoutebox}[2][]{%
  colframe=ngreen, 
  colback=ngreen,  
  left=2.75pt,                
  right=2.75pt,               
  boxsep=2pt,               
  #1
}

\newtcolorbox[auto counter]{definitionbox}[2][]{%
  colframe=nred,
  colback=nred,
  breakable,
  before upper={{\textbf{Definition \thetcbcounter.} #2}},
  #1
}

\makeatletter
\def\tcb@cnt@definitionboxautorefname{Def.}
\makeatother

\newtcolorbox[auto counter]{algbox}[2][]{%
  colframe=nred,
  colback=nred,
  breakable,
  left=1pt, right=1pt, top=0.5pt,
  breakable,
    before upper={{\textbf{Algorithm \thetcbcounter.} #2}},
  #1
}

\makeatletter
\def\tcb@cnt@algboxautorefname{Alg.}
\makeatother

\newcommand{\isd}[1]{}

\title{Interleaved Head Attention}

\author[2,\dagger, *]{Sai Surya Duvvuri}
\author[5, *]{Chanakya Ekbote}
\author[4, \dagger]{Rachit Bansal}
\author[3, \dagger]{Rishabh Tiwari}
\author[2, \dagger]{Devvrit Khatri}
\author[1]{David Brandfonbrener}
\author[5, \ddagger]{Paul Liang}
\author[2, \ddagger]{Inderjit Dhillon}
\author[1, \ddagger]{Manzil Zaheer}

\affiliation[1]{Meta}
\affiliation[2]{UT Austin}
\affiliation[3]{UC Berkeley}
\affiliation[4]{Harvard University}
\affiliation[5]{MIT}

\contribution[*]{Equal contribution}
\contribution[\ddagger]{Equal advising}
\contribution[\dagger]{Work done at FAIR/Meta}

\abstract{
    \vspace{-0.8em}
Multi-Head Attention (MHA) is the core computational primitive underlying modern Large Language Models (LLMs). However, MHA suffers from a fundamental \emph{linear scaling} limitation: $H$ attention heads produce exactly $H$ independent attention matrices, with no communication between heads during attention computation. This becomes problematic for multi-step reasoning, where correct answers depend on aggregating evidence from multiple parts of the context and composing latent token-to-token relations over a chain of intermediate inferences. To address this, we propose \emph{Interleaved Head Attention} (IHA), which enables cross-head mixing by constructing $P$ \emph{pseudo-heads} per head (typically $P=H$), where each pseudo query/key/value is a learned linear combination of all $H$ original queries, keys and values respectively. Interactions between pseudo-query and pseudo-key heads induce up to $P^2$ attention patterns per head with modest parameter overhead $\mathcal{O}(H^2P)$. We provide theory showing improved efficiency in terms of number of parameters on the synthetic Polynomial task (IHA uses $\Theta(\sqrt{k}n^2)$ parameters vs.\ $\Theta(kn^2)$ for MHA) and on the synthetic order-sensitive CPM-3 task (IHA uses $\lceil\sqrt{\nmax}\rceil$ heads vs.\ $\nmax$ for MHA). On real-world benchmarks, IHA improves Multi-Key retrieval on RULER by 10--20\% (4k--16k) and, after fine-tuning for reasoning on OpenThoughts, improves GSM8K by 5.8\% and MATH-500 by 2.8\% (Majority Vote) over full attention.
    \vspace{-0.8em}
}

\correspondence{\email{saisurya@cs.utexas.edu}, \email{cekbote@mit.edu}}

\begin{document}

\maketitle

\section{Introduction}

Multi-head attention (MHA) is the core computational primitive underlying modern Large Language Models \citep{vaswani2017attention}. In MHA, each of the $H$ heads computes an attention matrix independently: each query head attends only to its corresponding key--value projections, and heads do not interact during attention computation. Consequently, MHA exhibits a fundamental \emph{linear scaling} constraint: $H$ heads produce exactly $H$ attention matrices. While this is sufficient in many settings, it can be limiting for multi-step, compositional reasoning, where a token must aggregate evidence from multiple parts of the context and compose token-to-token relations over a chain of intermediate inferences. For example, in question answering, correct predictions often require chaining evidence: for example, to answer ``\emph{Where was the author of \textit{The Hobbit} born?}'', the model must first infer \textit{The Hobbit} $\rightarrow$ \textit{J.R.R.\ Tolkien} and then infer \textit{Tolkien} $\rightarrow$ \textit{born in South Africa}. This requires composing \emph{intermediate reasoning relations} rather than relying on a single direct association.

To formalize this limitation, we study the Polynomial Filter problem \citep{defferrard2016chebnet,chien2021gprgnn,lingam2021piecewisepolynomialfilteringapproach,ekbotefigure} as a controlled proxy for multi-step reasoning.
Concretely, an $r$-step dependency means that information reaches a token through $r$ successive relation applications (e.g., $r=1$ is direct, while $r=2$ composes two relations). We show that an individual MHA head (i.e., a single attention matrix) can represent only one such dependency pattern at a time.
Representing $k$ distinct chain lengths within one layer (i.e., simultaneously modeling direct and longer-range composed relations) requires $\Theta(k)$ heads (or additional depth), leading to parameter requirements that scale linearly with task complexity. Our theory on Polynomial Filters also generalize to other compositional primitives like binary relations \citep{cho2024strassen} and Match-3 \citep{sanford2023representational}.

In this work, we propose Interleaved Head Attention (IHA), which addresses this bottleneck.
Whereas standard MHA computes each head in isolation and therefore scales linearly with the number of distinct step-patterns it can represent, IHA breaks this constraint by enabling cross-head mixing within attention. For each head, IHA constructs $P$ pseudo-queries, pseudo-keys, and pseudo-values (typically $P=H$) as learned linear combinations of the original heads' query, key, and value projections. Interacting the $P$ pseudo-queries with the $P$ pseudo-keys induces up to $P^2$ attention patterns per head, yielding quadratic scaling in $P$ (and typically in $H$ when $P=H$). Unlike Talking-Heads and Knocking \citep{shazeer2020talking,zhou2025knocking}, which mix heads at the level of attention logits/weights, IHA performs this mixing before attention while preserving the standard attention operator, making it compatible with efficient kernels such as FlashAttention \citep{dao2022flashattention}. The extra parameters come only from pseudo-head mixing and scale as $\mathcal{O}(H^2P)$, which is modest since $H, P \ll d_{\text{model}}$, where $d_{\text{model}}$ denotes the model dimension. On Polynomial Filters of order $k$, we show that MHA needs $\Theta(k n^2)$ parameters, whereas IHA matches the same expressivity with $\Theta(\sqrt{k}n^2)$ parameters using $\mathcal{O}(\sqrt{k})$ heads.

We run extensive experiments on long-context modeling and supervised fine-tuning for reasoning. On the RULER long-context benchmark \citep{hsieh2024ruler}, IHA achieves 10--20\% relative improvements over full attention on Multi-Key Retrieval across 4k to 16k context lengths. After fine-tuning on OpenThoughts \citep{guha2025openthoughts} for reasoning, IHA outperforms full attention baselines by 5.8\% on GSM8K \citep{cobbe2021gsm8k} and 2.8\% on MATH-500 \citep{hendrycks2021math} under majority voting. Our contributions are:

\begin{mainbox}{\\}
 
\begin{enumerate}
\item \textbf{Interleaved Head Attention (IHA).} We introduce IHA, which enables cross-head mixing by constructing $P$ pseudo-queries, pseudo-keys, and pseudo-values per head (typically $P=H$); interactions between pseudo-queries and pseudo-keys induce up to $P^2$ attention patterns per head, rather than the $H$ independent attention matrices of standard MHA. \\

\item \textbf{Theory.} We prove that IHA is strictly more expressive than MHA while adding only \textbf{$4H^2P$} parameters. We also show improved asymptotic scaling on two synthetic benchmarks:  \textbf{$\Theta(\sqrt{k}\,n^2)$ vs.\ $\Theta(k n^2)$} parameters on Polynomial Filters, and \textbf{$\Theta(\nmax^{2.5})$ vs.\ $\Theta(\nmax^{3})$} attention cost on CPM-3 (best-known one-layer MHA). \\

\item \textbf{Empirical results.} Under FLOP-matched training, IHA improves Multi-Key retrieval on RULER by \textbf{10--20\%} (4k--16k) and, after OpenThoughts fine-tuning, improves GSM8K by \textbf{5.8\%} and MATH-500 by \textbf{2.8\%} (Maj@16) over full attention.
\end{enumerate}
\end{mainbox}

\section{Related Work}
\label{sec:related_work}

Recent work has begun to characterize when standard multi-head attention (MHA) is a poor fit for compositional, multi-step reasoning. A recurring theme is that many reasoning primitives require both (i) aggregating evidence spread across many positions and (ii) composing multiple intermediate transformations before producing an answer (e.g., relation/function composition and Match-3 \citep{kozachinskiy2025strassen,sanford2023representational}, as well as multi-hop QA benchmarks such as bAbI \citep{weston2016babi}). To study this behavior in a controlled theoretical fashion, we focus on two synthetic proxies: polynomial filters \citep{defferrard2016chebnet,chien2021gprgnn}, which model $k$-step aggregation, and CPM-3, which isolates order-sensitive composition and counting. In these settings, we show that IHA realizes the underlying primitives more efficiently than MHA, yielding quadratic improvements in the heads/parameters required by the corresponding constructions. In parallel, prior work extends attention along two main directions: adding richer token interactions or improving computational efficiency. Some methods go beyond standard pairwise query--key attention by modeling interactions among three or more tokens at once (e.g., simplicial/trilinear attention \citep{clift2019simplicial,roy2025fastsimplex}, Strassen-style constructions \citep{kozachinskiy2025strassen} and other multi-token mechanisms \citep{golovneva2025mta}) while increasing the number of parameters, while MQA/GQA \citep{shazeer2019mqa,ainslie2023gqa} reduce KV cost. Other lines of work mix information across heads (Talking-Heads, Knocking-Heads \citep{shazeer2020talking,zhou2025knocking}) or modify attention maps (e.g., Differential Attention \citep{ye2024differential}). IHA is complementary: it enables cross-head interaction within attention by mixing query, key and values into pseudo-heads, inducing quadratic interaction patterns while preserving the standard attention operator (and thus remaining compatible with FlashAttention \citep{dao2022flashattention}). An extended related works section appears in \autoref{app:ext_related_work}.

\section{Background}

\paragraph{Notation.}
We denote matrices with bold uppercase letters (e.g., $\bm{X}, \bm{W}$) and vectors with bold lowercase (e.g., $\bm{x}$).
For an input sequence of $N$ tokens with embedding dimension $D$, the input matrix is $\bm{X} \in \mathbb{R}^{N \times D}$.
We use $h$ to index attention heads, $d = D/H$ for per-head dimension where $H$ is the total number of heads,
and $\softmax(\cdot)$ for row-wise softmax. We use $[\bm{A}, \bm{B}]$ for column-wise (horizontal) concatenation and $[\bm{A}; \bm{B}]$ for row-wise (vertical) stacking.
In Algorithm~\ref{alg:iha}, we use \texttt{reshape}$(\bm{T}, [d_1, \ldots, d_k])$ to reshape tensor $\bm{T}$ to dimensions $d_1 \times \cdots \times d_k$, \texttt{einsum} for Einstein summation following NumPy conventions (e.g., \texttt{`mhp,nmd}$\to$\texttt{hpnd'} contracts index $m$ and permutes the result), and \texttt{merge\_pseudo} to interleave the pseudo-head dimension $P$ into the sequence dimension, transforming shape $(H,P,N,d) \to (H,NP,d)$. Throughout, $\mathbbm{1}[\cdot]$ denotes the indicator function; e.g., $\mathbbm{1}[x=y]=1$ if $x=y$ and $0$ otherwise. We also write this equivalently as $\bm{1}_{(x=y)}$.

\subsection{Multi-head Attention and Polynomial Filters}
\label{sec:multhead_attention}
\paragraph{MHA.} We use the standard scaled dot-product \emph{causal} self-attention mechanism \citep{vaswani2017attention}, with the
causal mask applied implicitly. 
\begin{definitionbox}[label=def:mha]
Given an input $\bm X\in\mathbb{R}^{N\times D}$ and $H$ heads with per-head dimension
$d\coloneqq D/H$, head $h\in[H]$ forms: $\bm Q^{(h)} \coloneqq \bm X\bm W_Q^{(h)}$, $
\ \bm K^{(h)} \coloneqq \bm X\bm W_K^{(h)},
\ \bm V^{(h)} \coloneqq \bm X\bm W_V^{(h)},
$ where $\bm W_Q^{(h)},\bm W_K^{(h)},\bm W_V^{(h)}\in\mathbb{R}^{D\times d}$ are learned projections. The per-head output is ($\forall h\in[H])$:
\begin{align*}
\widetilde{\bm X}^{(h)}
\;=\;
\softmax\!\left(\frac{1}{\sqrt d}\,\bm Q^{(h)}{\bm K^{(h)}}^\top\right)\bm V^{(h)}
\end{align*}
where the softmax is applied row-wise (and respects the causal mask). Finally, the head outputs are concatenated and
projected (with $\bm W_O\in\mathbb{R}^{D\times D}$ the output projection):
\begin{align*}
\widetilde{\bm X}
\;\coloneqq\;
\big[\widetilde{\bm X}^{(1)},\ldots,\widetilde{\bm X}^{(H)}\big]\bm W_O
\;\in\;\mathbb{R}^{N\times D},
\end{align*}
\end{definitionbox}

\paragraph{Polynomial filters.} Polynomial graph filters are a core primitive in graph signal processing. 
\begin{definitionbox}{(\textbf{Polynomial Graph Filters)}: } Given a graph
adjacency matrix $\bm{A} \in \mathbb{R}^{N \times N}$ and input features $\bm{X} \in \mathbb{R}^{N \times d}$,
one computes the polynomial filter bank,  $
\widetilde{\bm{X}} = \big[ \bm{X},\; \bm{A} \bm{X},\; \ldots,\; \bm{A}^{k-1}\bm{X} \big]$
\end{definitionbox}
The above defines a node representation which aggregates information from nodes up to $k$ hops away. This is a controlled proxy for
\emph{multi-step reasoning} in language tasks such as \cite{weston2016babi}: consider the story
\emph{(i)} ``Mary picked up the football.'' \emph{(ii)} ``Mary went to the kitchen.''
and the question ``Where is the football?'' A model must connect \emph{football} to \emph{Mary} (fact \emph{i})
and then \emph{Mary} to \emph{kitchen} (fact \emph{ii}), i.e., a two-step composition.
If we build a fact graph with one node per sentence and connect two nodes when they share an entity
(here, facts \emph{i} and \emph{ii} are connected via \emph{Mary}), then one-hop aggregation $\bm{A}\bm{X}$
retrieves the directly linked fact, while two-hop aggregation $\bm{A}^2\bm{X}$ captures precisely the
required two-step chain. In a language model, this graph is implicitly induced by the particular input
(and thus varies across examples); however, as a controlled proxy for analyzing multi-step information
propagation, we treat $\bm{A}$ as fixed for a given instance and study how well attention can realize
the corresponding $k$-hop operators. A natural question is therefore: how many attention heads does MHA need
to represent (or approximate) these $k$-hop aggregations, and to produce the full filter bank
$\big[\bm{X},\bm{A}\bm{X},\ldots,\bm{A}^{k-1}\bm{X}\big]$ in parallel? We aim to answer this question via \autoref{thm:mha_polyfilter_background}.

\begin{theorembox}[label=thm:mha_polyfilter_background]{\textbf{(MHA Polynomial Filter Representation):} }
Given a graph adjacency matrix $\bm{A} \in \mathbb{R}^{N \times N}$ and input features $\bm{X} \in \mathbb{R}^{N \times d}$, we concatenate the input with the identity matrix: $\widehat{\bm{X}} = [\bm{X}, \bm{I}]$. Then, representing the polynomial filter bank $\widetilde{\bm{X}} = [ \bm{X}, \bm{A}\bm{X}, \ldots, \bm{A}^{k-1}\bm{X} ]$ using one-layer linear MHA (without softmax) requires at least $k$ attention heads, with parameter complexity $2N(N+d)k + d(N+d)k$. Note that we assume $d\ll N$ and $k \ll N$.
\end{theorembox}

\paragraph{Proof sketch.} \textbf{Why does MHA need $k$ heads?}
We augment $\bm{X}$ with positional encodings $\widehat{\bm{X}}=[\bm{X},\bm{I}_N]$. In linear (no-softmax) attention, head $h$ outputs
\begin{align*}
\bm{O}_h
= \widehat{\bm{X}} \bm{W}_Q^{(h)} {\bm{W}_K^{(h)}}^\top \widehat{\bm{X}}^\top \widehat{\bm{X}} \bm{W}_V^{(h)}
\in \mathbb{R}^{N\times d}.
\end{align*}
To realize $\bm{A}^i\bm{X}$, choose
\begin{align*}
\bm{W}_Q^{(h)} &=
\begin{bmatrix}
\bm{0}_{d\times N} \\
\bm{A}^{i}
\end{bmatrix},
&
\bm{W}_K^{(h)} &=
\begin{bmatrix}
\bm{0}_{d\times N} \\
\bm{I}_N
\end{bmatrix},
&
\bm{W}_V^{(h)} &=
\begin{bmatrix}
\bm{I}_d \\
\bm{0}_{N\times d}
\end{bmatrix},
\end{align*}
which gives $\widehat{\bm{X}} \bm{W}_Q^{(h)} {\bm{W}_K^{(h)}}^\top \widehat{\bm{X}}^\top = \bm{A}^i$ and $\widehat{\bm{X}}\bm{W}_V^{(h)}=\bm{X}$, hence $\bm{O}_h=\bm{A}^i\bm{X}$. Since each head yields a single power $\bm{A}^i$, producing $\bm{A}^0,\ldots,\bm{A}^{k-1}$ requires at least $k$ heads. See \autoref{app:polynomialfilter} for more details. Note that the assumption that $d \ll N$ and $k \ll N$ is consistent with prior works on polynomial graph filters such as \cite{defferrard2016chebnet, chien2021gprgnn, lingam2021piecewisepolynomialfilteringapproach, ekbotefigure}.

\paragraph{Intuition.} The fundamental bottleneck in MHA is head isolation: head $h$ uses only its own query/key/value projections, yielding at most one attention pattern per head. Thus, if the target requires $k$ distinct relational patterns (e.g., $k$ polynomial terms), then a single MHA layer typically needs $\Omega(k)$ heads (or additional depth). IHA relaxes this constraint by constructing, for each head, $P$ pseudo-queries/ keys/ values as learned linear combinations of the original heads' query/ key/ value projections, so each head can realize up to $P^2$ attention patterns. In particular, setting $P=\lceil\sqrt{k}\rceil$ allows $k$ patterns to be realized within a single head, so IHA can represent $k$ patterns with only $H=\mathcal{O}(\sqrt{k})$ heads (up to constants/modeling constraints). We formalize this in \autoref{sec:iha}.

\section{Interleaved-Head Attention (IHA)}
\label{sec:iha}

\begin{figure}[h]
    \centering
    \includegraphics[width=\linewidth]{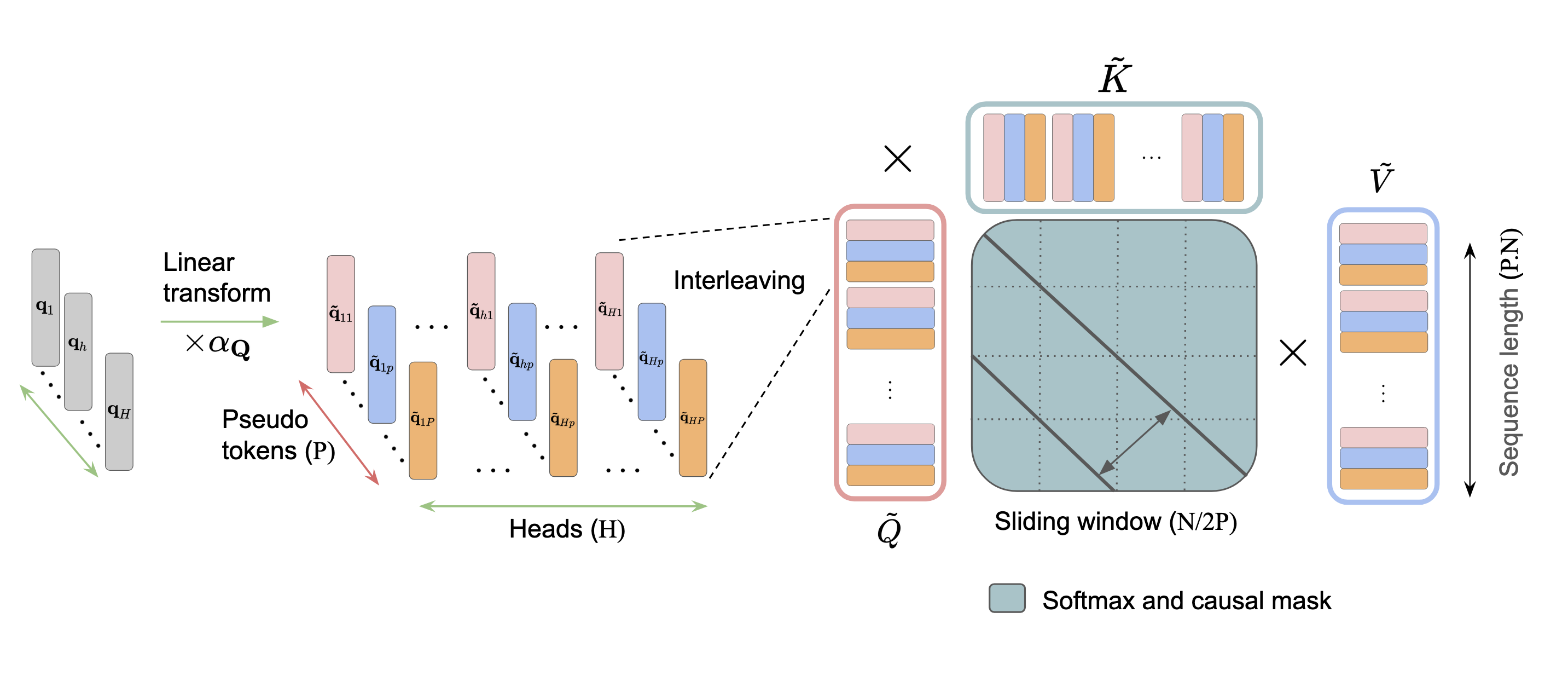}
    
    \caption{\textbf{Overview of Interleaved Head Attention (IHA).} 
    First, the model generates $P$ pseudo-tokens for each of the $H$ original heads via a learned linear transformation ($\times \mathbf{\mathcal{\alpha}_Q}$) operating on the heads axis (green). 
    These tokens are then interleaved to create an expanded sequence of length $P \cdot N$. 
    Finally, standard causal self-attention is computed on this expanded sequence, utilizing a sliding window (e.g., $N/2P$) to manage computational complexity while enabling cross-head interaction. Different linear transforms are used in query, key and values.}
    \label{fig:iha_architecture}
\end{figure}



IHA overcomes the one-to-one coupling of standard multi-head attention (MHA) by constructing, for each head, $P$ pseudo-queries, pseudo-keys, and pseudo-values as learned linear combinations of the $H$ queries, keys and values respectively (typically $P=H$). This enlarges the set of query/key/value projections and allows attention to mix information across heads, rather than restricting each query head to its paired key--value head. Within each head, the $P$ pseudo-queries attending to the $P$ pseudo-keys can induce up to $P^2$ distinct attention patterns, and this mechanism is applied independently across heads. The added expressivity incurs only modest overhead: pseudo-mixing weights scale as $\mathcal{O}(H^2P)$, which is small relative to the overall parameter budget since $H$ (and thus $P$) is typically much smaller than the model dimension. The full IHA algorithm is given in \autoref{alg:iha} and the architecture figure can be found in \autoref{fig:iha_architecture}.

\begin{algbox}[label=alg:iha]{\textbf{Interleaved-Head Attention} (\Cref{fig:iha_architecture})}

\begin{algorithmic}[1]
\Statex \textbf{Input:} $\bm{X}\in\mathbb{R}^{N\times D}$, heads $H$, pseudo-heads $P$, head dim $d=D/H$
\Statex \textbf{Params:} $\bm{W}_Q, \bm{W}_K, \bm{W}_V, \bm{W}_O \in \mathbb{R}^{D \times D}$; $\bm{\alpha}^{Q,K,V} \in \mathbb{R}^{H \times H \times P}$; $\bm{R} \in \mathbb{R}^{H \times P}$
\Statex \textbf{Output:} $\widetilde{\bm{X}} \in \mathbb{R}^{N \times D}$

\Statex \textbf{// Step 1: Project and reshape to per-head representations}
\State $\bm{Q}, \bm{K}, \bm{V} \gets \texttt{reshape}(\bm{X}\bm{W}_Q, \bm{X}\bm{W}_K, \bm{X}\bm{W}_V)$ \Comment{$\to [N,H,d]$}

\Statex \textbf{// Step 2: Generate pseudo-heads via learned mixing across heads}
\State $\widetilde{\bm{Q}} \gets \texttt{einsum}(\texttt{`mhp,nmd}\to\texttt{hpnd'}, \bm{\alpha}^Q, \bm{Q})$ \Comment{$(H,P,N,d)$}
\State $\widetilde{\bm{K}} \gets \texttt{einsum}(\texttt{`mhp,nmd}\to\texttt{hpnd'}, \bm{\alpha}^K, \bm{K})$ \Comment{$(H,P,N,d)$}
\State $\widetilde{\bm{V}} \gets \texttt{einsum}(\texttt{`mhp,nmd}\to\texttt{hpnd'}, \bm{\alpha}^V, \bm{V})$ \Comment{$(H,P,N,d)$}

\Statex \textbf{// Step 3: Merge pseudo dimension into sequence dimension (interleaved)}
\State $\overline{\bm{Q}}, \overline{\bm{K}}, \overline{\bm{V}} \gets \texttt{merge\_pseudo}(\widetilde{\bm{Q}}, \widetilde{\bm{K}}, \widetilde{\bm{V}})$ \Comment{$\to [H,NP,d]$}

\Statex \textbf{// Step 4: Standard scaled dot-product attention per head}
\State $\overline{\bm{O}}_h \gets \softmax\left(\frac{1}{\sqrt{d}} \overline{\bm{Q}}_h \overline{\bm{K}}_h^\top\right) \overline{\bm{V}}_h, \quad \forall h$
\Comment{$\overline{\bm{O}} \in \mathbb{R}^{H \times NP \times d}$}

\Statex \textbf{// Step 5: Unmerge and collapse pseudo-heads}
\State $\bm{P} \gets \texttt{reshape}(\overline{\bm{O}}, \texttt{[H,N,P,d]})$ \Comment{Reshape to $(H,N,P,d)$}
\State $\bm{O} \gets \texttt{einsum}(\texttt{`hp,hnpd}\to\texttt{hnd'}, \bm{R}, \bm{P})$ \Comment{Collapse: $(H,N,d)$}

\Statex \textbf{// Step 6: Concatenate heads and project output}
\State $\widetilde{\bm{X}} \gets \texttt{reshape}(\bm{O}, \texttt{[N,D]}) \cdot \bm{W}_O$

\State \Return $\widetilde{\bm{X}}$
\end{algorithmic}
\end{algbox}

In \autoref{alg:iha}, Step~2 constructs, for each head $h$ and token index $n\in[N]$, $P$ pseudo-head tokens by taking learned linear combinations of the $H$ original heads’ query, key and value projections. Step~3 interleaves them by replacing each original token with $P$ consecutive virtual tokens, so the sequence becomes $(1,1),(1,2),\ldots,(1,P),(2,1),\ldots,(N,P)$ where $(n,p)$ denotes the $p$-th pseudo-head token at position $n$. Step~4 then runs standard scaled dot-product attention once on this length-$NP$ sequence. This lets different pseudo-head tokens attend differently (including to different pseudo-head tokens at the same original position), yielding up to $P^2$ attention patterns per head (and $HP^2$ overall) without custom kernels. Interleaving is useful with RoPE because RoPE depends on the position index: giving each $(n,p)$ its own virtual position assigns each pseudo-head token a distinct RoPE phase, and variable-length inference is handled by generating RoPE for length $NP$. The procedure is also compatible with FlashAttention~\citep{dao2022flashattention}, since Step~4 is standard attention. For the theoretical analysis, \autoref{def:IHA} gives an equivalent, more algebraic formulation of IHA that omits interleaving and the output projection. Since our proofs do not use positional encodings (e.g., RoPE), interleaving is unnecessary, and dropping the projection cleanly isolates the core pseudo-head mixing and attention computation.

\begin{definitionbox}[label=def:IHA]{\textbf{Interleaved Head Attention}}
\begin{algorithmic}
\Require $\bm{X}\in\mathbb{R}^{N\times D}$, heads $H$, pseudos $P$, per-head dim $d$
\Require $\bm{W}_Q^{(m)},\bm{W}_K^{(m)},\bm{W}_V^{(m)}\in\mathbb{R}^{D\times d}$ for $m=1,\ldots,H$
\Require $\alpha^Q,\alpha^K,\alpha^V\in\mathbb{R}^{H\times H\times P}$ and $\bm{R}\in\mathbb{R}^{H\times HP}$

\Statex \textbf{Pseudo-head mixing across heads}
\For{$h=1$ to $H$}
  \For{$j=1$ to $P$}
    \State $\widetilde{\bm{Q}}_{h,j} \coloneqq \sum_{m=1}^{H}\alpha^Q_{m,h,j}\,\bm{X} \bm{W}_Q^{(m)} \in\mathbb{R}^{N\times d}$
    \State $\widetilde{\bm{K}}_{h,j} \coloneqq \sum_{m=1}^{H}\alpha^K_{m,h,j}\,\bm{X} \bm{W}_K^{(m)} \in\mathbb{R}^{N\times d}$
    \State $\widetilde{\bm{V}}_{h,j} \coloneqq \sum_{m=1}^{H}\alpha^V_{m,h,j}\,\bm{X} \bm{W}_V^{(m)} \in\mathbb{R}^{N\times d}$
  \EndFor
\EndFor

\Statex \textbf{Pseudo-major stacking (row-wise concatenation to length $PN$)}
\For{$h=1$ to $H$}
  \State $\overline{\bm{Q}}_h \coloneqq \left[\widetilde{\bm{Q}}_{h,1}^\top; \ldots; \widetilde{\bm{Q}}_{h,P}^\top\right]^\top \in\mathbb{R}^{PN\times d}$
  \State $\overline{\bm{K}}_h \coloneqq \left[\widetilde{\bm{K}}_{h,1}^\top; \ldots; \widetilde{\bm{K}}_{h,P}^\top\right]^\top \in\mathbb{R}^{PN\times d}$
  \State $\overline{\bm{V}}_h \coloneqq \left[\widetilde{\bm{V}}_{h,1}^\top; \ldots; \widetilde{\bm{V}}_{h,P}^\top\right]^\top \in\mathbb{R}^{PN\times d}$
\EndFor

\Statex \textbf{Attention (per head)}
\For{$h=1$ to $H$}
  \State $\bm{S}_h \coloneqq \frac{1}{\sqrt{d}}\,\overline{\bm{Q}}_h\,\overline{\bm{K}}_h^\top \in\mathbb{R}^{PN\times PN}$
  \State $\overline{\bm{P}}_h \coloneqq \softmax(\bm{S}_h)\,\overline{\bm{V}}_h\in\mathbb{R}^{PN\times d}$
\EndFor

\Statex \textbf{Unstack and collapse $HP\to H$}
\For{$h=1$ to $H$}
  \For{$t=1$ to $N$}
    \State $\bm{O}_h[t,:] \coloneqq \sum_{h'=1}^{H}\sum_{j=1}^{P} \bm{R}_{h,(h'-1)P+j}\; \bm{P}_{h',j}[t,:]\in\mathbb{R}^{d}$
  \EndFor
\EndFor

\Statex \textbf{Concatenate heads}
\State $\widetilde{\bm{X}} \coloneqq \left[\bm{O}_1, \bm{O}_2, \ldots, \bm{O}_H\right]\in\mathbb{R}^{N\times D}$
\end{algorithmic}
\end{definitionbox}

In the following sections, we (i) establish a strict expressivity separation by showing that the class of functions representable by MHA is contained in (and generally a strict subset of) those representable by IHA, (ii) analyze IHA on two synthetic benchmarks (the polynomial filter and CPM3 tasks; defined later), and (iii) experimentally show that IHA outperforms other attention variants.

\subsection{IHA Strictly Generalizes MHA}
\label{sec:iha_superset}

We formalize the sense in which IHA strictly generalizes standard multi-head attention (MHA) while making the parameter overhead explicit. Fix a sequence length $N$ and number of heads $H$. Let $\mathcal{M}$ denote the set of all single-layer $H$-head MHA modules with query, key, value matrices as defined in \autoref{def:mha}, requiring $Q$ parameters in total. Let $\mathcal{P}_{P}$ denote the corresponding set of $H$-head IHA modules as in \autoref{alg:iha} (and \autoref{def:IHA}) with $P$ pseudo-heads per head, whose  query, key, and value weight matrices have the same dimensions as those in MHA (hence also contributing $Q$ parameters), but which additionally introduce mixing tensors $\alpha^Q,\alpha^K,\alpha^V\in\mathbb{R}^{H\times H\times P}$ and a collapse map $\bm{R}\in\mathbb{R}^{H\times HP}$.

\begin{theorembox}[label=thm:iha_superset_params_main]
\textbf{(IHA Superset Property).}  For any $P\ge 1$, every module in $\mathcal{P}_{P}$ has $Q+4H^2P$ parameters (namely $Q$ from the query, key, value projections plus $3H^2P$ from $\alpha^Q,\alpha^K,\alpha^V$ and $H^2P$ from $\bm{R}$). Moreover, for every $P\ge 1$, $\mathcal{M}\subseteq \mathcal{P}_{P}$ and for every $P\ge 2$ the inclusion is strict: $\mathcal{M}\subsetneq \mathcal{P}_{P}.$ 
\end{theorembox}

\noindent\textbf{Proof sketch.}
\emph{Inclusion.} Fix any MHA instance with weights $\{\bm{W}_Q^{(m)},\bm{W}_K^{(m)},\bm{W}_V^{(m)}\}_{m=1}^H$. We construct an IHA instance (with any chosen $P\ge 1$) that computes the same function by selecting parameters that ignore all but one pseudo-channel. Specifically, for all $m,i\in[H]$ and all $j\in[P]$, set $
\alpha^Q_{m,i,j}=\bm{1}_{(m=i)},\ 
\alpha^K_{m,i,j}=\bm{1}_{(m=i)},\ \alpha^V_{m,i,j}=\bm{1}_{(m=i)}$
and choose $\bm{R}$ to select only the $(i,1)$ pseudo-block:
\begin{align*}
\bm{R}_{\,i,\,(i'-1)P+j}=\begin{cases}
1 & \text{if } i'=i \text{ and } j=1,\\
0 & \text{otherwise.}
\end{cases}
\end{align*}
Then $\widetilde{\bm{Q}}_{i,j}=\bm{X}\bm{W}_Q^{(i)}$, $\widetilde{\bm{K}}_{i,j}=\bm{X}\bm{W}_K^{(i)}$, and $\widetilde{\bm{V}}_{i,j}=\bm{X}\bm{W}_V^{(i)}$ for all $j$, so the stacked attention produces copies of the original MHA head outputs and the collapse map returns exactly the MHA outputs. Hence $\mathcal{M}\subseteq\mathcal{P}_{P}$.

\emph{Strictness.} Consider the repeated-token subspace
$\mathcal{S}=\{\bm{X}=\bm{1}_N\bm{x}^\top:\bm{x}\in\mathbb{R}^d\}, \ N\ge 2 $. On $\mathcal{S}$, every MHA head has identical queries/keys/values at all positions, so each score matrix has identical rows and the row-wise softmax is uniform; consequently each head output reduces to $\bm{1}_N\bm{x}^\top\bm{W}_V^{(m)}$, which is linear in $\bm{x}$. Therefore every $H$-head MHA module is linear on $\mathcal{S}$. In contrast, IHA with $P=2$ can be parameterized (using only the additional $4H^2P$ mixing/collapse parameters on top of the same  projections) so that the stacked attention produces a nonlinear function of $\bm{x}$ on $\mathcal{S}$: for example, by creating two pseudo-query/key variants with opposite signs and choosing the pseudo-values/collapse so that the output involves a difference of softmax-normalized terms depending on the cosine score $\langle \bm{x}^\top\bm{W}_Q, \bm{x}^\top\bm{W}_K\rangle$, yielding a nonlinearity (e.g., a $\tanh$-like dependence). Since no MHA can be nonlinear on $\mathcal{S}$, this IHA mapping cannot be represented by any MHA, proving $\mathcal{M}\subsetneq\mathcal{P}_{P}$ for all $P\ge 2$. For a detailed proof please refer to \autoref{app:ihasuperset}

\subsection{Representing Polynomial Filters using IHA}
\label{sec:polyfilter}

As established in \autoref{sec:multhead_attention}, polynomial graph filters provide a clean proxy for multi-hop information propagation (and thus multi-step reasoning): the $i$-th term $\bm{A}^i\bm{X}$ aggregates information from $i$-hop neighborhoods. Computing $\bm{X}, \bm{A}\bm{X}, \ldots, \bm{A}^{k-1}\bm{X}$ in parallel captures $k$-step composition in a controlled setting. In \autoref{thm:polyfilter}, we show how IHA represents this polynomial filter.

\begin{theorembox}[label=thm:polyfilter]
\textbf{(Representing Polynomial Filters).}
Given a graph adjacency matrix $\bm{A}\in\mathbb{R}^{N\times N}$ and input features $\bm{X}\in\mathbb{R}^{N\times d}$ with $d<N$, we concatenate the input with the identity matrix $\widehat{\bm{X}}=[\bm{X},\bm{I}]$. For one-layer attention-based multi-head architectures \emph{without softmax} that are capable of representing all polynomial filter constructions with $k$ heads, there exists an equivalent one-layer attention-based \iha~architecture without softmax that requires only $\lceil\sqrt{k}\rceil$ heads. In terms of parameter complexity, an \mha~construction with $k$ heads requires $2N(N{+}d)k + d(N{+}d)k$ parameters, whereas the equivalent \iha~construction with $\lceil\sqrt{k}\rceil$ heads requires $2N(N{+}d)\lceil\sqrt{k}\rceil + d(N{+}d)\lceil\sqrt{k}\rceil^{2} + 4\lceil\sqrt{k}\rceil^{3}$ parameters. Here $d$ denotes the embedding dimension, $k$ the polynomial order, and $N$ the number of nodes, and we assume $k\ll N$ and $d\ll N$.
\end{theorembox}

\paragraph{Proof sketch.}
We consider representing the polynomial filter bank, $\widetilde{\bm{X}}
\;=\;
\big[\bm{X},\;\bm{A}\bm{X},\;\ldots,\;\bm{A}^{k-1}\bm{X}\big]$ using a single \emph{linear-attention} layer (i.e., without softmax). Since $\bm{X}\in\mathbb{R}^{N\times d}$ is low-rank when $d<N$, we augment the input with positional encodings $\widehat{\bm{X}}=[\bm{X},\bm{I}]$.

\textbf{Why MHA needs $k$ heads.}
As established in \Secref{sec:multhead_attention}, in linear MHA, each head produces exactly one attention operator (matrix) $\bm{S}_h\in\mathbb{R}^{N\times N}$. To represent all $k$ distinct powers $\bm{A}^0,\ldots,\bm{A}^{k-1}$ in parallel, one therefore needs $k$ independently parameterized heads. The parameter-minimal exact MHA construction thus uses $k$ heads.

\textbf{Why IHA only needs $\lceil\sqrt{k}\rceil$ heads.}
Let $H=\lceil\sqrt{k}\rceil$ (and in the construction we set the number of pseudo-heads to $P=H$). IHA exploits the factorization
\begin{align*}
\bm{A}^{i}
\;=\;
\bm{A}^{(h-1)H+(j-1)},
\qquad
h,j\in\{1,\ldots,H\},
\end{align*}
so that $H^2\ge k$ distinct powers can be generated via pairwise query--key interactions. Instead of assigning one head per power, IHA assigns heads to \emph{blocks} of powers. Concretely, we choose $H$ query and key matrices (where $\forall \ h, j\  \in\  \{1, \cdots, H\} $
\begin{align*}
\bm{W}_{Q,\texttt{IHA}}^{(h)}
\;=\;
\begin{bmatrix}
\bm{0}_{d\times N}\\
\bm{A}^{(h-1)H}
\end{bmatrix} \qquad \bm{W}_{K,\texttt{IHA}}^{(j)}
\;=\;
\begin{bmatrix}
\bm{0}_{d\times N}\\
\pth{\bm{A}^{j-1}}^\top
\end{bmatrix},
\end{align*}
When query head $h$ interacts with key head $j$, the resulting (linear) attention matrix is
\begin{align*}
\bm{S}_{h,j}
&=
\widehat{\bm{X}}\bm{W}_{Q,\texttt{IHA}}^{(h)}
\big(\bm{W}_{K,\texttt{IHA}}^{(j)}\big)^{\!\top}
\widehat{\bm{X}}^{\!\top}
=
\bm{A}^{(h-1)H+(j-1)}.
\end{align*}
Thus, $H$ query matrices and $H$ key matrices generate $H^2\ge k$ distinct polynomial powers through pairwise interaction. Pseudo-head mixing ensures that, within each head $h$, a \emph{single} query attends to \emph{all} $H$ key/value branches, producing the entire block
\begin{align*}
\big[
\bm{A}^{(h-1)H}\bm{X},\;
\bm{A}^{(h-1)H+1}\bm{X},\;
\ldots,\;
\bm{A}^{(h-1)H+(H-1)}\bm{X}
\big]
\end{align*}
in one shot. Value matrices are chosen to route each $\bm{A}^{(h-1)H+(j-1)}\bm{X}$ into a distinct $d$-dimensional output block, and concatenating the $H$ heads recovers $\big[\bm{X},\;\bm{A}\bm{X},\;\ldots,\;\bm{A}^{H^2-1}\bm{X}\big]$ with any extra $(H^2-k)$ blocks treated as padding. Consequently, IHA represents the same polynomial filter bank using only $O(\sqrt{k})$ heads, reducing the dominant parameter cost from $\Theta(kN^2)$ (MHA) to $\Theta(\lceil\sqrt{k}\rceil N^2)$, up to lower-order routing terms (including the $4H^3$ pseudo-mixing/collapse parameters). For more details and the full proof, see \autoref{app:polynomialfilter}.

\subsection{\mbox{Representing CPM-3 using IHA}}
\label{sec:cpm3}


Polynomial filters provide a controlled proxy for multi-step retrieval: they aggregate information from $k$-hop neighborhoods, but the result is essentially a bag of $k$-hop evidence. To probe a complementary regime, we introduce \emph{Count Permutation Match-3} (CPM-3), which isolates order-sensitive composition and counting. For each position $i$, the model ranges over ordered pairs of other positions $(j_1,j_2)$, checks whether the triple $(x_i,x_{j_1},x_{j_2})$ satisfies a simple modular predicate, and outputs how many ordered pairs satisfy it. CPM-3 is an arithmetic analogue of multi-fact QA \cite{weston2016babi}: instead of only retrieving relevant facts, the model must combine two facts in the correct order and then count how many such combinations exist. For example, in a story with facts of the form ``$u$ is in $v$,'' a query about $u=i$ asks how many ordered pairs of facts $(j_1,j_2)$ form a valid two-step chain, the first fact says $i$ is in some $z$, and the second fact says that same $z$ is in some $y$. Each ordered pair that correctly links through a shared intermediate $z$ is a valid supporting pair for the query, and the answer is the count of all such pairs. We formalize this intuition by encoding tokens as scalars and replacing this chain test with the modular relation defined below.

\paragraph{Count Permutation Match-3 (CPM-3).}
We introduce the CPM-3 task. The input is a length-$N$ sequence of natural numbers $(x_1,\ldots,x_N)$, with $N\le \nmax$. For each position $i$, the desired output is
\begin{align*}
\mathrm{CPM}_i(3)
\;=\;
\bigl|\{(j_1,j_2)\in[N]^2:\ \phi(x_i,x_{j_1},x_{j_2})=0\}\bigr|,
\end{align*}
where the predicate is the (order-sensitive) modular expression $\phi(x_i,x_{j_1},x_{j_2})
\;:=\;
x_i + Gx_{j_1} + x_{j_2}\ \mathrm{mod}\ M,$ with modulus $M\in\mathbb{N}$ and coefficient $G>2M$. The condition $G>2M$ ensures the predicate is not permutation invariant: typically $\phi(x_i,x_{j_1},x_{j_2})\neq \phi(x_i,x_{j_2},x_{j_1})$ when $x_{j_1}\neq x_{j_2}$. We next ask how efficiently different attention mechanisms can realize CPM-3 in a single layer; in particular, we show that \iha~admits a construction with $\ceil{\sqrt{\nmax}}$ heads, whereas known \mha~constructions require $\nmax$ heads (in \autoref{thm:permutationmatch3}).

\begin{theorembox}[label=thm:permutationmatch3]
\textbf{(Count Permutation Match-3).}
Let $\nmax$ denote the maximum number of tokens that can be processed by the model in the worst case. There exists a one-layer transformer with \emph{interleaved-head attention} (\iha) that can represent CPM-3 using $\ceil{\sqrt{\nmax}}$ attention heads. The number of parameters required by this IHA construction is upper bounded by $37 \nmax^2 \sqrt{\nmax} + \nmax^2(\nmax-1) + \nmax^2$. In contrast, the best currently known construction based on \emph{multi-head attention} (\mha) requires $\nmax$ attention heads, and its parameter count is lower bounded by $3 \nmax^3 + \nmax^2(\nmax-1) + \nmax^2$. Throughout, we assume the vocabulary size is at most on the order of the maximum sequence length, i.e., $|\mathcal{V}|=O(\nmax)$.
\end{theorembox}

\paragraph{Proof sketch.}
We sketch why CPM-3 can be implemented with $\ceil{\sqrt{\nmax}}$ IHA heads but (in known constructions) requires $\nmax$ MHA heads. The task is to output, for each position $i$, the count of ordered pairs $(j_1,j_2)$ such that
\begin{align*}
x_i + Gx_{j_1} + x_{j_2} \equiv 0\quad (\mathrm{mod}\ M),
\end{align*}
with $G>2M$ ensuring order sensitivity. As in the polynomial-filter construction, we use positional encodings to make positions addressable:
\begin{align*}
\widehat{\bm{X}}=[\bm{X},\bm{I}]\in\mathbb{R}^{\nmax\times(\nmax+1)},
\qquad
\bm{X}\in\mathbb{R}^{\nmax\times 1}.
\end{align*}
Note that the CPM-3 output at position $i$ depends on \emph{all} ordered pairs $(j_1,j_2)$, so a convenient one-layer strategy is to first use attention to build, at every $i$, a local ``workspace'' that contains all token values $\{x_j\}_{j=1}^{\nmax}$ in a fixed, known order. A downstream MLP can then (i) select any ordered pair of coordinates $(j_1,j_2)$, (ii) form $x_i+Gx_{j_1}+x_{j_2}$, (iii) test the modulo constraint, and (iv) sum indicators. We now sketch why MHA needs $\nmax$ heads to build this workspace, while IHA needs only $\ceil{\sqrt{\nmax}}$.

\textbf{Why MHA needs $\nmax$ heads.}
With positional encodings $\hat{\bm X}=[\bm X,\bm I]$ and hard attention (softmax temperature $0$), each MHA head can implement one cyclic shift of the sequence. Let $\bm P\in\mathbb R^{\nmax\times\nmax}$ be the cyclic permutation matrix. For $h\in\{1,\ldots,\nmax\}$ set $\bm{W}_{V,\texttt{MHA}}^{(h)}=
\begin{bmatrix}
1  & \bm{0}_{\nmax\times 1}^\top
\end{bmatrix}^\top$ and, 
\begin{align*}
\bm{W}_{Q,\texttt{MHA}}^{(h)}&=
\begin{bmatrix}
\bm{0}_{1\times\nmax}\\
\bm{P}^{h-1}
\end{bmatrix},
\quad
\bm{W}_{K,\texttt{MHA}}^{(h)}=
\begin{bmatrix}
\bm{0}_{1\times\nmax}\\
\bm{I}_{\nmax\times\nmax} 
\end{bmatrix}
\end{align*}
and so the head produces $\bm S_h=\bm P^{h-1}$ and outputs $\bm P^{h-1}\bm X$. Concatenating all heads yields
$\big[\bm X,\bm P\bm X,\ldots,\bm P^{\nmax-1}\bm X\big]$, 
which places all $\nmax$ symbols into each position's workspace. Since each head contributes only one shift $\bm P^t$, producing all $\nmax$ shifts in one layer requires $\nmax$ heads, giving attention-parameter scaling $\Omega(\nmax^3)$.

\textbf{Why IHA only needs $\ceil{\sqrt{\nmax}}$ heads.}
Let $H=\ceil{\sqrt{\nmax}}$ and set $P=H$. IHA factors each shift index $t\in\{0,\ldots,\nmax-1\}$ as
\begin{align*}
t=(h-1)H+(j-1),
\qquad h,j\in\{1,\ldots,H\},
\end{align*}
so $\bm P^{t}=\bm P^{(h-1)H}\bm P^{j-1}$. Define query and key matrices by
\begin{align*}
\bm{W}_{Q,\texttt{IHA}}^{(h)}=
\begin{bmatrix}
\bm{0}_{1\times\nmax}\\
\bm{P}^{(h-1)H}
\end{bmatrix},
\qquad
\bm{W}_{K,\texttt{IHA}}^{(j)}=
\begin{bmatrix}
\bm{0}_{1\times\nmax}\\
\big(\bm{P}^{j-1}\big)^\top
\end{bmatrix},
\end{align*}
so the $(h,j)$ interaction realizes $\bm S_{h,j}=\bm P^{(h-1)H+(j-1)}$. The crucial difference from MHA is that pseudo-head mixing lets a \emph{single} query head $h$ attend to \emph{all} $j\in\{1,\ldots,H\}$ key/value heads, producing in one head the block
\begin{align*}
\big[
\bm P^{(h-1)H}\bm X,\;
\bm P^{(h-1)H+1}\bm X,\;
\ldots,\;
\bm P^{(h-1)H+(H-1)}\bm X
\big],
\end{align*}
with value projections routing each shift into a distinct coordinate block. Concatenating over $h\in\{1,\ldots,H\}$ yields all $\{\bm P^t\bm X\}_{t=0}^{\nmax-1}$ at each position. i.e., the same workspace as above, but using only $H=\ceil{\sqrt{\nmax}}$ heads. The downstream MLP is then identical to that of the MHA construction. The key difference is that IHA can realize many distinct attention patterns per head: with $P$ pseudo-queries and $P$ pseudo-keys per head, each head can implement up to $P^2$ different attention maps, and across $H$ heads this gives up to $HP^2$ patterns. Taking $P=H=\ceil{\sqrt{\nmax}}$ provides enough distinct patterns to cover the $\nmax$ required shifts while reducing the head count from $\nmax$ to $\ceil{\sqrt{\nmax}}$. Consequently, the best-known one-layer MHA construction incurs $\Theta(\nmax^{3})$ attention cost, whereas the IHA construction achieves $\Theta(\nmax^{2}\sqrt{\nmax})$ (up to the $O(H^3)$ pseudo-mixing/collapse overhead). For full details, see \autoref{app:cpm3}.

\section{Experiments}

\newcommand{\deltag}[2]{\cellcolor{green!#2}{\textbf{+}#1}}
\newcommand{\deltar}[2]{\cellcolor{red!#2}{\textbf{--}#1}}
\newcommand{\deltaz}{\cellcolor{gray!10}{--}}
\newcommand{\oursrow}{\rowcolor{blue!6}} 

We evaluate Interleaved Head Attention (IHA) in large-scale language model training to answer two questions: (i) does IHA improve long-context retrieval and length generalization when adapting models beyond their pretraining window, and (ii) does IHA improve reasoning on math and code benchmarks before and after supervised fine-tuning? To isolate architectural effects, we keep the backbone, optimizer, data, and training budget fixed across variants and compare against strong attention baselines. Building on prior sections showing that IHA is strictly more expressive than standard MHA and yields separations on controlled reasoning proxies, we test whether these advantages translate into practical gains during pretraining, long-context adaptation, and downstream evaluation. We also report additional experiments on synthetic reasoning datasets in \autoref{app:synthetic_reasoning}.

\subsection{Experimental Setup}

\paragraph{Model architecture.}
All experiments use a 2.4B-parameter decoder-only Transformer with hidden size 2560, 26 layers, and $H=20$ attention heads (head dimension 128). We use a $4\times$ FFN expansion (FFN size 10{,}240), vocabulary size 128{,}256 (Llama~3 tokenizer; \citep{dubey2024llama3}), and RoPE positional encoding \citep{su2021roformer} with $\theta=500{,}000$. Pretraining context length is 8{,}192 tokens.

\paragraph{Training.}
All models are trained for 240{,}000 steps (240B tokens) with identical hyperparameters: peak learning rate $8\times 10^{-4}$ with 1{,}000-step warmup and cosine decay \citep{loshchilov2017sgdr} to $8\times 10^{-6}$, AdamW \citep{loshchilov2019adamw} ($\beta_1{=}0.9$, $\beta_2{=}0.95$, weight decay 0.1), gradient clipping 1.0, and BF16 mixed precision \citep{micikevicius2018mixed}. Training uses FSDP \citep{zhao2023fsdp} over 128 H200 GPUs.

\paragraph{Baselines.}
We compare five attention mechanisms. (1) \textbf{Global Attention}~\citep{vaswani2017attention} is standard multi-head self-attention, where every layer attends to the full sequence. (2) \textbf{Global+Local}~\citep{vaswani2017attention} is a hybrid schedule that alternates local sliding-window attention (window size 512) with periodic global-attention layers in a 4:1 ratio. (3) \textbf{Talking Heads}~\citep{shazeer2020talking} augments multi-head attention by learning to mix information across heads both before and after the softmax, enabling richer head-to-head interactions. (4) \textbf{Diff Transformer}~\citep{ye2024differential} defines attention as the difference of two softmax attention maps, which can sharpen or suppress patterns via contrastive weighting. and (5)  \textbf{IHA (Ours)}, interleaved-head attention with pseudo-heads. Let $N$ be the sequence length, $d$ the per-head dimension, $H$ the number of heads, and $P$ the number of pseudo-heads per head. Since interleaving expands the effective sequence length from $N$ to $NP$, global IHA has per-head complexity $O((NP)^2 d)=O(P^2N^2d)$, i.e., a factor-$P^2$ over global MHA; we therefore FLOP-match all comparisons. We use a hybrid local--global schedule (four sliding-window IHA layers with $W \coloneqq N/(2P^2)$ followed by one global layer) so the average cost matches the global-attention baseline up to constants; see \autoref{app:compute} for details.	\looseness=-2

\paragraph{Benchmarks.}
For long-context evaluation we use RULER \citep{hsieh2024ruler}. For reasoning and coding we evaluate GSM8K \citep{cobbe2021gsm8k}, MATH-500 \citep{hendrycks2021math}, MBPP \citep{austin2021mbpp}, and HumanEval \citep{chen2021humaneval}.

\begin{table*}[t]
    \centering
    \caption{SFT evaluation after fine-tuning on OpenThoughts. IHA achieves the best overall performance, with larger gains over the baselines than at pre-training. $\Delta$ is relative to \textbf{Global Attention} (green/red denote improvement/regression), and \textbf{Avg.\ Rank}$\downarrow$ is the mean rank across metrics (lower is better).}
    \label{tab:sft-eval}
    \footnotesize
    \setlength{\tabcolsep}{3pt}
    \resizebox{1\textwidth}{!}{%
    \begin{tabular}{l cc cc cc cc cc cc cc cc cc c}
        \toprule
        \textbf{Model}
        & \textbf{GSM8K P@1} & \textbf{$\Delta$}
        & \textbf{GSM8K Maj@16} & \textbf{$\Delta$}
        & \textbf{MATH-500 P@1} & \textbf{$\Delta$}
        & \textbf{MATH-500 Maj@16} & \textbf{$\Delta$}
        & \textbf{MBPP P@1} & \textbf{$\Delta$}
        & \textbf{MBPP P@10} & \textbf{$\Delta$}
        & \textbf{Avg.\ Rank$\downarrow$} \\
        \midrule
        \oursrow
        \textbf{IHA (Ours)}
        & \textbf{34.3\%} & \deltag{4.8}{65}
        & \textbf{54.2\%} & \deltag{5.8}{70}
        & \textbf{10.0\%} & \deltag{1.2}{30}
        & \textbf{18.4\%} & \deltag{2.8}{45}
        & 15.5\% & \deltag{0.8}{25}
        & 41.6\% & \deltag{0.4}{20}
        & \textbf{1.5} \\
        \midrule
        \textbf{Global Attention}
        & 29.5\% & \deltaz
        & 48.4\% & \deltaz
        & 8.8\% & \deltaz
        & 15.6\% & \deltaz
        & 14.7\% & \deltaz
        & 41.2\% & \deltaz
        & 3.8 \\
        \textbf{Global+Local}
        & 26.5\% & \deltar{3.0}{45}
        & 46.9\% & \deltar{1.5}{30}
        & 7.6\% & \deltar{1.2}{30}
        & 15.0\% & \deltar{0.6}{22}
        & 15.0\% & \deltag{0.3}{18}
        & 41.9\% & \deltag{0.7}{24}
        & 4.3 \\
        \textbf{Talking Heads}
        & 29.3\% & \deltar{0.2}{20}
        & 49.4\% & \deltag{1.0}{25}
        & 7.8\% & \deltar{1.0}{25}
        & 18.2\% & \deltag{2.6}{42}
        & \textbf{15.9\%} & \deltag{1.2}{30}
        & \textbf{43.1\%} & \deltag{1.9}{35}
        & 2.5 \\
        \textbf{Diff Transformer}
        & 31.6\% & \deltag{2.1}{38}
        & 53.5\% & \deltag{5.1}{65}
        & 9.0\% & \deltag{0.2}{18}
        & 18.0\% & \deltag{2.4}{40}
        & 15.3\% & \deltag{0.6}{22}
        & 39.2\% & \deltar{2.0}{36}
        & 2.8 \\
        \bottomrule
    \end{tabular}%
    }
\end{table*}

\begin{table*}[h]
    \centering
    \caption{Pre-trained model evaluation (5-shot). IHA achieves the best overall reasoning performance, improving over \textbf{Global Attention} and \textbf{Global+Local} on GSM8K. $\Delta$ is relative to \textbf{Global Attention} (green/red denote improvement/regression), and \textbf{Avg.\ Rank}$\downarrow$ is the mean rank across reported metrics (lower is better).}
    \label{tab:pretrain-eval}
    \footnotesize
    \setlength{\tabcolsep}{3pt}
    \resizebox{1\textwidth}{!}{%
    \begin{tabular}{l cc cc cc cc cc c}
        \toprule
        \textbf{Model}
        & \textbf{GSM8K EM} & \textbf{$\Delta$}
        & \textbf{GSM8K Maj@5} & \textbf{$\Delta$}
        & \textbf{MATH-500 EM} & \textbf{$\Delta$}
        & \textbf{MBPP P@1} & \textbf{$\Delta$}
        & \textbf{HumanEval P@1} & \textbf{$\Delta$}
        & \textbf{Avg.\ Rank$\downarrow$} \\
        \midrule
        \oursrow
        \textbf{IHA (Ours)}
        & \textbf{8.34\%} & \deltag{2.73}{60}
        & \textbf{8.42\%} & \deltag{2.81}{62}
        & \textbf{3.54\%} & \deltag{0.66}{22}
        & 24.5\% & \deltag{1.1}{28}
        & 17.1\% & \deltar{0.1}{18}
        & \textbf{1.4} \\
        \midrule
        \textbf{Global Attention}
        & 5.61\% & \deltaz
        & 5.61\% & \deltaz
        & 2.88\% & \deltaz
        & 23.4\% & \deltaz
        & \textbf{17.2\%} & \deltaz
        & 2.9 \\
        \textbf{Global+Local}
        & 6.82\% & \deltag{1.21}{38}
        & 6.90\% & \deltag{1.29}{40}
        & 2.26\% & \deltar{0.62}{22}
        & 23.6\% & \deltag{0.2}{18}
        & 16.0\% & \deltar{1.2}{28}
        & 2.9 \\
        \textbf{Talking Heads}
        & 5.46\% & \deltar{0.15}{18}
        & 5.38\% & \deltar{0.23}{20}
        & -- & --
        & 23.8\% & \deltag{0.4}{20}
        & 16.0\% & \deltar{1.2}{28}
        & 4.0 \\
        \textbf{Diff Transformer}
        & 5.46\% & \deltar{0.15}{18}
        & 5.61\% & \deltaz
        & -- & --
        & \textbf{25.0\%} & \deltag{1.6}{32}
        & 15.4\% & \deltar{1.8}{35}
        & 3.5 \\
        \bottomrule
    \end{tabular}%
    }
\end{table*}

\subsection{Long Context Evaluation}

\begin{figure}[t]
\centering
\includegraphics[width=\columnwidth]{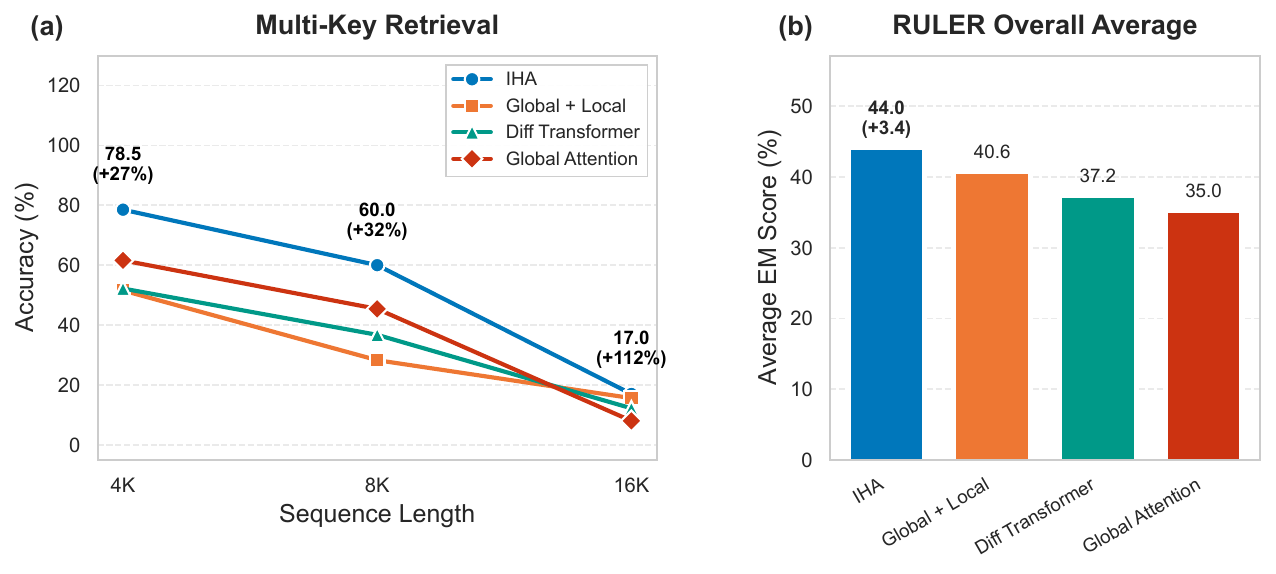}
\caption{\textbf{RULER long-context results after 64k fine-tuning.}
(a) Multi-Key Retrieval accuracy at 4k/8k/16k context lengths (orange: IHA improvement over Sliding Window).
(b) Overall RULER Exact Match (EM) show strong improvements using IHA.}
\label{fig:ruler_eval}
\end{figure}

For long-context evaluation, we fine-tuned all models at 64k (beyond the pretraining window) and evaluated on RULER (\autoref{fig:ruler_eval}). IHA is consistently stronger on retrieval: on Multi-Key Retrieval it improves over Global Attention by \textbf{+27\%} (4k), \textbf{+32\%} (8k), and \textbf{+112\%} (16k). Across the full RULER suite, IHA achieves the best average  EM (exact match) (\textbf{44.0\%}), outperforming Global+Local (\textbf{40.6\%}), Diff Transformer (\textbf{37.2\%}), and Global Attention (\textbf{35.0\%}).

\subsection{Reasoning Evaluation}


We evaluate \emph{pre-trained} models in a 5-shot setting to probe reasoning ability prior to supervised instruction tuning (\autoref{tab:pretrain-eval}). \textbf{IHA (Ours)} consistently improves over \textbf{Global Attention} on the core reasoning benchmarks: on GSM8K it achieves the best scores (\textbf{8.34\%} EM and \textbf{8.42\%} Maj@5; \textbf{+2.73}/\textbf{+2.81}), and it also leads on MATH-500 EM with \textbf{3.54\%} (\textbf{+0.66}). Coding results are mixed, with a modest gain on the MBPP benchmark to 24.5\% (second best) while HumanEval is near parity, but IHA is the most consistent method overall, achieving the best mean rank across reported metrics (\textbf{Avg.\ Rank}$\downarrow=1.4$). Overall, these results indicate that IHA’s added expressivity translates into stronger reasoning performance even before downstream fine-tuning.


\paragraph{Supervised fine-tuning.}
We fine-tuned all variants on OpenThoughts \citep{guha2025openthoughts} (8B tokens) and evaluated with temperature 0.6 using 16 generations (\autoref{tab:sft-eval}). \textbf{IHA (Ours)} achieves the best overall performance, leading on all reasoning metrics (e.g., \textbf{54.2\%} GSM8K Maj@16, \textbf{+5.8} over \textbf{Global Attention}; \textbf{18.4\%} MATH-500 Maj@16, \textbf{+2.8}) . On coding (MBPP), \textbf{Talking Heads} is best (15.9\% P@1, 43.1\% P@10), while IHA is \emph{second best} on both MBPP metrics (15.5\% P@1, 41.6\% P@10) suggesting that while IHA's expressivity excels at logical state tracking for math, head mixing is well-suited for function-level code generation. Overall, IHA remains consistently strong across tasks after SFT, whereas other variants tend to peak on specific datasets. and attaining the best \textbf{Avg.\ Rank}$\downarrow$ (\textbf{1.5})
)

\section{Conclusion}
We introduced \emph{Interleaved Head Attention (IHA)}, which overcomes MHA's linear scaling by learning $P$ pseudo-queries, pseudo-keys, and pseudo-values per head as linear combinations of the original heads. Interactions between pseudo-queries and pseudo-keys induce up to $P^2$ attention patterns per head. Our theory shows improved parameter efficiency on Polynomial Filters (IHA uses $\Theta(\sqrt{k}n^2)$ parameters vs.\ $\Theta(kn^2)$ for MHA) and on the order-sensitive CPM-3 task (IHA uses $\lceil\sqrt{\nmax}\rceil$ heads vs.\ $\nmax$ for MHA). Empirically, under FLOP-matched training, IHA improves Multi-Key retrieval on RULER by 10--20\% (4k--16k) and, after OpenThoughts fine-tuning, improves reasoning on GSM8K by 5.8\% and MATH-500 by 2.8\% (majority vote) over full attention. \textbf{Limitations.} Global IHA can increase attention cost (scaling as $O(P^2N^2)$), which we mitigate with a sliding-window schedule; future work includes adaptive pseudo-head allocation and extensions to encoder--decoder and vision architectures.

\section*{Acknowledgements}

We thank Rohan Anil for their comments on the IHA algorithm and Niladri S. Chatterji for helping with the experiment setup. 


\clearpage

\bibliography{section/references}
\bibliographystyle{plainnat}

\newpage

\appendix
\onecolumn

\vspace{10em}
\begin{center}
\rule{\textwidth}{1pt}\\[0.75em]
{\LARGE\bfseries Appendix}\\[0.5em]
\rule{\textwidth}{1pt}
\end{center}

\tableofcontents

\newpage

\section{Extended Related Work}
\label{app:ext_related_work}

\paragraph{Hardness results for compositional reasoning.}
Recent theory has begun to formalize when standard multi-head attention (MHA) is an inefficient mechanism for compositional multi-step reasoning. \citet{kozachinskiy2025strassen} prove hardness results for binary relation composition and function composition, and \citet{sanford2023representational} show that Match-3 requires $\Theta(N^3)$ parameters under standard attention-based constructions. These settings share two structural requirements: global aggregation of evidence across many positions, and composition of intermediate relational signals prior to producing an output. Related behavior also appears in multi-hop QA benchmarks such as bAbI \citep{weston2016babi}, which require combining information across multiple hops. We study these challenges through two controlled proxies that separate these requirements. Polynomial filters provide a standard $k$-hop aggregation primitive from spectral GNNs \citep{defferrard2016chebnet,chien2021gprgnn, lingam2021piecewisepolynomialfilteringapproach, ekbotefigure}, and CPM-3 isolates order-sensitive composition and counting. In both settings, we show that IHA realizes the relevant primitives with quadratic improvements in head and parameter efficiency relative to comparable MHA constructions.

\paragraph{Higher-order and multi-token attention.}
Several works enrich token interactions by going beyond pairwise query-key attention. The 2-Simplicial Transformer \citep{clift2019simplicial} generalizes attention to trilinear interactions, and \citet{roy2025fastsimplex} provide an efficient Triton implementation. Strassen-style attention constructions use fast matrix multiplication ideas to accelerate particular compositional patterns \citep{kozachinskiy2025strassen}. Multi-Token Attention \citep{golovneva2025mta} introduces mechanisms that mix information across small token groups, for example via local mixing over attention weights. These approaches typically modify the attention operator or introduce specialized higher-order structure. IHA is complementary. It preserves the standard attention operator and induces effective higher-order behavior by learning cross-head mixing of $Q$, $K$, and $V$ into pseudo-heads, yielding a broad family of quadratic interaction patterns while remaining compatible with optimized attention kernels.

\paragraph{Iterative computation via depth or recurrence.}
Another approach to multi-step reasoning is to increase the number of sequential transformations, either via depth or recurrence. Looped transformers \citep{saunshi2025looped} show that iterating a $k$-layer block for $L$ loops can match a $kL$-layer model, and can support chain-of-thought-like behavior through repeated refinement. These methods increase computation across iterations. Our approach is complementary. IHA increases within-layer interaction capacity by constructing $P$ pseudo-heads per head, typically with $P=H$, using learned linear combinations of the original queries, keys, and values. Interactions between pseudo queries and pseudo keys induce up to $P^2$ attention patterns per head within a single attention computation, without adding sequential depth.

\paragraph{Efficient and cross-head attention variants.}
A large body of work improves attention efficiency and head utilization. Multi-Query Attention (MQA) \citep{shazeer2019mqa} and Grouped Query Attention (GQA) \citep{ainslie2023gqa} reduce inference cost by sharing key and value projections across heads. Other methods explicitly couple heads by mixing attention logits or weights, including Talking-Heads \citep{shazeer2020talking} and Knocking-Heads \citep{zhou2025knocking}, or by shaping attention maps, as in Differential Attention \citep{ye2024differential}. In contrast, IHA enables cross-head interaction within attention by mixing $Q$, $K$, and $V$ representations into pseudo-heads via learned linear combinations. This induces quadratic interaction structure while preserving the standard attention computation, and it remains compatible with efficient kernels such as FlashAttention \citep{dao2022flashattention}.

\section{Theoretical Properties of IHA}
\label{app:theorotical_properties}

In this section, we establish theoretical properties of IHA that highlight its advantages over MHA. We present and prove several key representational results below.

\subsection{IHA Superset Property}
\label{app:ihasuperset}

\begin{theorembox}[label=thm:pseudoiha_superset_params]
{(IHA Superset Property; Parameter-Aware): }
Fix a sequence length $n$ and number of heads $h$.
Let $\mathcal{M}$ denote the set of all single-layer $h$-head multi-head attention (MHA) modules with query, key, value matrices and inputs bounded in the Frobenius norm, requiring $Q$ parameters in total.

Let $\mathcal{P}_{p}$ denote the corresponding set of $h$-head IHA modules as in \autoref{alg:iha} with $p$ pseudos per head, whose base query, key, and value weight matrices have the same dimensions as those in MHA (hence also contributing $Q$ parameters), but which additionally introduce
\[
\alpha^Q,\alpha^K,\alpha^V\in\mathbb{R}^{h\times h\times p}
\quad\text{and}\quad
R^\ell\in\mathbb{R}^{h\times hp},
\]
i.e., an additional $3h^2p + h^2p = 4h^2p$ parameters, for a total of $Q+4h^2p$ parameters.

Then for every $p\ge 1$,
\begin{align*}
    \mathcal{M} \subseteq \mathcal{P}_{p},
\end{align*}
and for every $p\ge 2$ the inclusion is strict:
\begin{align*}
    \mathcal{M} \subsetneq \mathcal{P}_{p}.
\end{align*}
\end{theorembox}

\begin{proof}
\textbf{(Inclusion $\mathcal{M}\subseteq \mathcal{P}_{p}$).}
Fix any \mha~instance with weights $\{W_Q^{(m)},W_K^{(m)},W_V^{(m)}\}_{m=1}^h$.
We construct a IHA instance (with any chosen $p\ge 1$) that realizes the same function by selecting parameters that ignore all but one pseudo-channel.

Specifically, set for all $m,i\in\{1,\ldots,h\}$ and $\forall \ j \  \in \  \{1, \cdots, p\} $,
\begin{align*}
\alpha^Q_{m,i,j}=\bm{1}_{(m=i)},\qquad
\alpha^K_{m,i,j}=\bm{1}_{(m=i)},\qquad
\alpha^V_{m,i,j}=\bm{1}_{(m=i)},
\end{align*}

Further, choosing $R^\ell\in\mathbb{R}^{h\times hp}$ so that it selects only the $(i,1)$ pseudo-block:
\begin{align*}
R^\ell_{\,i,\,(i'-1)p+j}=\begin{cases}
1 & \text{if } i'=i \text{ and } j=1,\\
0 & \text{otherwise.}
\end{cases}
\end{align*}
Then for each head $i$ we have $\widetilde{\bm{Q}}_{i,j}=\bm{X}\bm{W}_Q^{(i)}$, $\widetilde{\bm{K}}_{i,j}=\bm{X}\bm{W}_K^{(i)}$, $\widetilde{\bm{V}}_{i,j}=\bm{X}\bm{W}_V^{(i)}$ $ \forall j \in \{1, \cdots, p \}$.
Consequently, the stacked attention produces an output whose only nonzero contribution is exactly the usual MHA head output, and the collapse via $R^\ell$ returns $O_i$ equal to that MHA head output. Concatenating heads yields exactly the original MHA module. Hence $\mathcal{M}\subseteq\mathcal{P}_{p}$.

\medskip
\textbf{Strictness for $p\ge 2$ (works for any $n\ge 2$ and any $d$ and any $h\ge 1$).}
It suffices to exhibit one $h$-head IHA/PseudoIHA configuration (with $p\ge 2$) that cannot be represented by any $h$-head MHA configuration.

\medskip
\textbf{MHA is linear on repeated-token inputs.}
Fix $n\ge 2$ and consider inputs with repeated tokens
\begin{align*}
\bm{X}=\mathbf{1}_n \bm{x}^\top \in \mathbb{R}^{n\times d},
\end{align*}
where $\mathbf{1}_n\in\mathbb{R}^n$ is the all-ones vector and $\bm{x}\in\mathbb{R}^d$.
For any MHA head $m$, define
\begin{align*}
\bm{q}:=\bm{x}^\top \bm{W}_Q^{(m)},\qquad \bm{k}:=\bm{x}^\top \bm{W}_K^{(m)},\qquad \bm{v}:=\bm{x}^\top \bm{W}_V^{(m)}.
\end{align*}
Then every token position has identical query/key/value, so the attention score matrix is constant:
\begin{align*}
\bm{S}^{(m)} = (\bm{X}\bm{W}_Q^m)(\bm{X}\bm{W}_K^m)^\top
= (\mathbf{1}_n \bm{q})(\mathbf{1}_n \bm{k})^\top
= (\langle \bm{q},\bm{k}\rangle)\,\mathbf{1}_n\mathbf{1}_n^\top.
\end{align*}
Since each row of $S^{(m)}$ is the same, hence the row-wise softmax is uniform:
\begin{align*}
\sigma\!\big(S^{(m)}\big)=\frac{1}{n}\mathbf{1}_n\mathbf{1}_n^\top.
\end{align*}
Therefore the head output equals
\begin{align*}
\mathrm{Att}(\bm{X}\bm{W}_Q^{(m)},\bm{X}\bm{W}_K^{(m)},\bm{X}\bm{W}_V^{(m)})
=\frac{1}{n}\mathbf{1}_n\mathbf{1}_n^\top(\mathbf{1}_n v)
=\mathbf{1}_n \bm{v}
=\mathbf{1}_n \bm{x}^\top \bm{W}_V^{(m)},
\end{align*}
which is linear in $\bm{x}$. Concatenating heads and applying any fixed output projection preserves linearity in each head.
Hence \emph{every} $h$-head \mha~module is linear on the repeated-token subspace
$\{\mathbf{1}_n \bm{x}^\top: \bm{x}\in\mathbb{R}^d\}$.

\medskip
\textbf{IHA yields a non-linear mapping of the input when $p\ge 2$.}
It is enough to consider the case $p=2$, since for any $p>2$ we can deactivate the additional pseudo channels by setting their mixing coefficients to zero. Concretely, we could impose
\begin{align*}
\alpha^{Q}_{m,i,j}=\alpha^{K}_{m,i,j}=\alpha^{V}_{m,i,j}=0
\qquad \forall\, m,i\in\{1,\dots,h\},\ \forall\, j>2,
\end{align*}
so that only the first two pseudo heads contribute (with the right reduction matrix $R$). We now construct a \iha~layer whose output is \emph{nonlinear} even on repeated-token inputs of the form $\bm{X}=\mathbf{1}_n \bm{x}^\top$.

Towards this, we focus on only one head $h$ with two psudo tokens / heads. We let $\alpha^{Q}_{m,i,1}=\alpha^{K}_{m,i,1}=\alpha^{V}_{m,i,1} = \alpha^{V}_{m,i,2} = \bm{1}_{(m=i)}$ and $\alpha^{Q}_{m,i,2}=\alpha^{K}_{m,i,2} = -\bm{1}_{(m=i)}$. Hence, for any arbitrary head $h$, we obtain,
\begin{align*}
    \overline{\bm{Q}}_h = \begin{bmatrix}
        \widetilde{\bm{Q}}_{h,1} \\ \widetilde{\bm{Q}}_{h,2}
    \end{bmatrix} \qquad \overline{\bm{K}}_h = \begin{bmatrix}
        \widetilde{\bm{K}}_{h,1} \\ \widetilde{\bm{K}}_{h,2}
    \end{bmatrix} \qquad \overline{\bm{V}}_h = \begin{bmatrix}
        \widetilde{\bm{V}}_{h,1} \\ \widetilde{\bm{V}}_{h,2}
    \end{bmatrix}
\end{align*}
Hence, on computing, we obtain:
\begin{align*}
    \overline{\bm{Q}}_h = \begin{bmatrix}
    \bm{Q}_{h}\\ -\bm{Q}_{h}
    \end{bmatrix} \qquad \overline{\bm{K}}_h = \begin{bmatrix}
       {\bm{K}}_{h} \\ - {\bm{K}}_{h}
    \end{bmatrix} \qquad \overline{\bm{V}}_h = \begin{bmatrix}
        {\bm{V}}_{h} \\ {\bm{V}}_{h}
    \end{bmatrix}
\end{align*}
Note, that here, $\bm{Q}_{h}, \bm{K}_{h}, \bm{V}_{h}$ refer to the query, key and value matrices of \mha~respectively. Hence, on computing attention, we obtain:
\begin{align*}
    \overline{\bm{P}}_h &= \text{softmax}\pth{\begin{bmatrix}
        \bm{Q}_h \bm{K}_h^\top & -\bm{Q}_h \bm{K}_h^\top \\
        -\bm{Q}_h \bm{K}_h^\top & \bm{Q}_h \bm{K}_h^\top
    \end{bmatrix}} \begin{bmatrix}
        \bm{V}_{h} \\ \bm{V}_{h}
    \end{bmatrix}
\end{align*}
Further, choosing $R^\ell\in\mathbb{R}^{h\times hp}$ so that it selects only the $(i,1)$ pseudo-block:
\begin{align*}
R^\ell_{\,i,\,(i'-1)p+j}=\begin{cases}
1 & \text{if } i'=i \text{ and } j=1,\\
0 & \text{otherwise.}
\end{cases}
\end{align*}
The output of each head is:
\begin{align*}
    \bm{O}_h = \text{softmax}\pth{\begin{bmatrix}
        \bm{Q}_h \bm{K}_h^\top & -\bm{Q}_h \bm{K}_h^\top \\
    \end{bmatrix}} \begin{bmatrix}
        \bm{V}_{h} \\ \bm{V}_h
    \end{bmatrix}
\end{align*}

Thus the overall \iha~layer computes a nonlinear function of the input data.

\medskip
\textbf{Conclusion.}
On the repeated-token subspace $\{\mathbf{1}_n \bm{x}^\top : \bm{x}\in\mathbb{R}^d\}$, every $h$-head \mha~layer reduces to a linear map in $\bm{x}$, whereas the $h$-head \iha~construction above is nonlinear on the same set. Consequently, no $h$-head \mha~configuration can represent this \iha~mapping, and thus $\mathcal{M}\subsetneq \mathcal{P}_p$ for all $p\ge 2$.
\end{proof}

\subsection{Representing Polynomial Filters}
\label{app:polynomialfilter}
\begin{theorembox}[label=thm:polyfilter_proof]
{(Representing Polynomial Filters): } Given a graph adjacency matrix $\bm{A} \in \mathbb{R}^{N \times N}$ and input features $\bm{X} \in \mathbb{R}^{N \times d}$ with $d < N$, we concatenate the input with the identity matrix: $\widehat{\bm{X}} = [\bm{X}, \bm{I}]$. For one-layer attention-based multi-head architectures without softmax that are capable of representing all polynomial filter constructions with $k$ heads, there exists an equivalent one-layer attention-based \iha~architecture without softmax that requires only $\lceil \sqrt{k} \rceil$ heads. In terms of parameter complexity, an \mha~construction with $k$ heads requires $2n(N+d)k + d(N+d)k$ parameters, whereas the equivalent \iha~construction with $\lceil \sqrt{k} \rceil$ heads requires $2n(N+d)\lceil \sqrt{k} \rceil + d(d+N)\lceil \sqrt{k} \rceil^2 + 4\ceil{\sqrt{k}}^3$ parameters. Here, $d$ denotes the embedding dimension, $k$ the number of hops in the polynomial filter, and $N$ the number of nodes, and we assume that $k << N$ and $d<<N$.
\end{theorembox}

\paragraph{Background.} 
Given fixed input data $\bm{X} \in \mathbb{R}^{N \times d}$ and a full-rank graph adjacency matrix $\bm{A} \in \mathbb{R}^{N \times N}$, 
the goal of this task is to obtain representations that depend on the graph in a polynomial manner. 
Specifically, given a polynomial order $k$, our objective is to compute:
\begin{align*}
    \widetilde{\bm{X}} \coloneqq [\bm{X}, \bm{A}\bm{X}, \ldots, \bm{A}^{k-1}\bm{X}].
\end{align*} Note that, in most cases, it is not possible to recover $\bm{A}$ or its powers $\bm{A}^i$ from any linear combination of the input features $\bm{X}$, primarily because $\bm{X}$ is typically sparse or low-rank. 

\begin{proof}
In this proof, our goal is to determine whether we can represent \(\widetilde{\bm{X}}\) using MHA and IHA and if yes, the goal is to understand the number of parameters needed to do so. 
Since \(\bm{A}\) is full-rank, we cannot directly use \(\bm{X}\). 
Instead, we define 
\[
\widehat{\bm{X}} = 
\begin{bmatrix}
\bm{X} & \bm{I}
\end{bmatrix},
\]
which effectively augments \(\bm{X}\) with positional encodings. 
Henceforth, we work with \(\widehat{\bm{X}}\). 
We now examine the constructions and parameter requirements for MHA and IHA, omitting explicit layer dependence \(\ell\) for brevity.

\paragraph{MHA.} 
From the definition, for linear attention, each head in MHA computes
\begin{align*}
    \widetilde{\bm{X}}^{(h)}_{\texttt{MHA}} 
    &= \left(\widehat{\bm{X}}\bm{W}_{Q,\texttt{MHA}}^{h}
        \left(\bm{W}_{K,\texttt{MHA}}^{h}\right)^{\!\top} \widehat{\bm{X}}^{\!\top}\right)
        \widehat{\bm{X}}\bm{W}_{V,\texttt{MHA}}^{h}.
\end{align*}
Since \(d < N\), we have 
\(\widehat{\bm{X}}\bm{W}_{Q,\texttt{MHA}}^{h}
        \left(\bm{W}_{K,\texttt{MHA}}^{h}\right)^{\!\top}
        \widehat{\bm{X}}^{\!\top} \in \mathbb{R}^{N \times N}\).
To recover a polynomial filter, we must be able to solve
\begin{align*}
    [\bm{X}, \bm{A}\bm{X}, \ldots, \bm{A}^{k-1}\bm{X}]
    =
    \left[\widetilde{\bm{X}}^{(1)}_{\texttt{MHA}}, \ldots, \widetilde{\bm{X}}^{(H)}_{\texttt{MHA}}\right].
\end{align*}
There are multiple design choices that can satisfy this equation. 
However, we are constrained by the fact that 
\(\bm{W}_{V,\texttt{MHA}}^{h}\) cannot depend on the input data \(\widehat{\bm{X}}\), 
and that the downstream embedding dimension must be \(kd\). 
Under these constraints, we can construct a minimal solution such that, for each head \(h\),
\begin{align*}
    \widehat{\bm{X}}\bm{W}_{V,\texttt{MHA}}^{h} &= \bm{X}, \\
    \left(\widehat{\bm{X}}\bm{W}_{Q,\texttt{MHA}}^{h}\left(\bm{W}_{K,\texttt{MHA}}^{h}\right)^{\!\top}\right) \widehat{\bm{X}}^{\!\top} &= \bm{A}^{h-1}.
\end{align*}
This construction requires exactly $2n(N+d)k + d(N+d)k $ parameters. To show that it is indeed minimal, we proceed as follows. We first argue that atleast $k$ heads are needed to represent $\widetilde{\bm{X}}$, and then we make arguments about the parameters. Towards this we argue about the rank of $\widetilde{\bm{X}}$. By definition, we know that:
\begin{align*}
    \widetilde{\bm{X}} = \begin{bmatrix}
        \bm{X} & \bm{A}\bm{X} & \cdots & \bm{A}^{k-1}\bm{X}
    \end{bmatrix}
\end{align*}
Hence, 
\begin{equation}
\label{eq:polynomial_filter_rank}
\begin{aligned}
    \text{rank}(\widetilde{\bm{X}}) &= \text{min}\pth{N, \sum_{i=0}^k \text{rank}(\bm{A}^i \bm{X})} \\
    &\leq \text{min}(N, kd) \\
    &\leq kd
\end{aligned}
\end{equation}
Where in the above, we have used that $\text{rank}(\bm{A}^i \bm{X}) \leq d, \forall\ i \in \{0,\ \cdots,\ k-1\}$, that $kd < N$ by assumption. Moreover, we would also like to note that for a generic $\bm{X}$, the rank can be tight. That is $\exists X$ such that $\text{rank}(\widetilde{\bm{X}}) = kd$. It is tight. Hence we will make the argument that ecven IHA needs to have ranks greater or equal to this to be able to represnt the output. Hence, towards this lets first assume that we have the number of heads $H$ to be less than $k$. Moreover, to match the dimensions of $\widetilde{\bm{X}}$, we let the value weights for some heads to be arbitrary such that the final dimensions match. We weould like to note that for any particular head $h$ \(\bm{W}_{K,\texttt{MHA}}^{h}\) must depend only on \(\bm{A}\) and not on \(\bm{X}\); hence, we can write
\begin{align*}
\bm{W}_{K,\texttt{MHA}}^{h} = 
\begin{bmatrix}
    B_h(\bm{A})_{d \times d_h}  \\ 
    C_h(\bm{A})_{N \times d_h}
\end{bmatrix}.
\end{align*}
Note that for the dimensions to match, we would need $\sum_{h=1}^{H}d_h = kd$. Moreover, per head, we define for shorthand, $\text{Att}_h(\widehat{\bm{X}}) \coloneqq \left(\widehat{\bm{X}}\bm{W}_{Q,\texttt{MHA}}^{h}\left(\bm{W}_{K,\texttt{MHA}}^{h}\right)^{\!\top}\right) \widehat{\bm{X}}^{\!\top}$. 

We let the representation of MHA for $H$ number of heads less than $k$ be denoted by $\widetilde{\bm{X}'}$, where:
\begin{align*}
    \widetilde{\bm{X}'} &= \begin{bmatrix}
        \text{Att}_1(\widehat{\bm{X})}\bm{X}B_1(\bm{A}) + \text{Att}_1(\widehat{\bm{X})}C_1(\bm{A}),\ \cdots,\ \text{Att}_h(\widehat{\bm{X})}\bm{X}B_H(\bm{A}) + \text{Att}_H(\widehat{\bm{X})}C_H(\bm{A})
    \end{bmatrix}
\end{align*}
Now since we want $\widetilde{\bm{X}'} = \widetilde{\bm{X}}, \forall~ \bm{X}$, if we substitute $\bm{X}=\bm{0}$, then we can easily see that would imply that $\forall h \in \{1, \cdots, H\}~ \text{Att}_hC_h(\bm{A}) = \bm{0}$. However, $\text{Att}_h \neq 0$ as if it were $\bm{0}$, then $\widetilde{\bm{X}'} = \bm{0}$ which would then imply that $\widetilde{\bm{X}'} \neq \widetilde{\bm{X}}$ for any arbitrary non-zero $\bm{X}$. Hence, the only solution is that $C_h(\bm{A}) =\bm{0}$. Hence, we can cleanly write that:
\begin{align*}
    \widetilde{\bm{X}'} &= \begin{bmatrix}
        \text{Att}_1(\widehat{\bm{X})}\bm{X}B_1(\bm{A}),\ \cdots,\ \text{Att}_H(\widehat{\bm{X})}\bm{X}B_H(\bm{A})
    \end{bmatrix}
\end{align*}
Now we try to compute the rank of $\widetilde{\bm{X}'}$. Note that again, we have assumed that $H<k$. Hence, 
\begin{equation}
\begin{aligned}
    \text{rank}(\widetilde{\bm{X}'}) &= \text{min}\pth{N, \sum_{i=1}^H \text{rank}(\text{Att}_1(\widehat{\bm{X})}\bm{X}B_i(\bm{A}))} \\
    &\leq \text{min}\pth{N, \sum_{i=1}^H \text{min}\pth{\text{rank}(\text{Att}_1(\widehat{\bm{X})}),\  \text{rank}(\bm{X}),\  \text{rank}(B_i(\bm{A})))}} \\
    &\leq \text{min}(N, Hd) \\
    &\leq Hd
\end{aligned}
\end{equation}
Now from \autoref{eq:polynomial_filter_rank}, we can see that the rank of the concatenation of the embeddings of polynomial filter are $kd$ when tight, however the rank in the case of  $\widetilde{\bm{X}'}$ is at most $Hd$. If $H<k$, clearly the polynomial filter is more expressive and has a higher rank than that of  $\widetilde{\bm{X}'}$ which implies that if $H<k$, then the polynomial filter cannot be represented Multi-Head attention. Now we argue that for heads $H>k$, there are constructions that can represnt the polynomial filter, however, it is not tight in terms of the parameters. Let us assume that there are $H>k$ heads. Moreover, we only make the arguement for the first head of MHA, and then this argument will essentially hold for all of the heads. We split this situation into multiple cases. Note that we are not really specifying what the dimension of each head in this case and assume it to be arbitrary. Hence, for the first case, we assume that the output dimension of the head is less than that of the first output of the polynomial filter. Hence, for equality, we need:
\begin{align*}
    \text{Att}_1(\widehat{\bm{X}})XB_1(\bm{A}) = XQ_1
\end{align*}
Note that $B_1(\bm{A})\in \mathbb{R}^{N\times d_1}$, where $d_1 \leq d$ Note that $Q_1 \in \mathbb{R}^{d \times d_1}$, and clearly $Q_1$ is a $d_1$ rank matrix with one hot vectors across each column to isolate the rows that correspond to $\text{Att}_1(\widehat{\bm{X}})XB_1(\bm{A})$. Therefore, on taking the $\text{vec}(\cdot)$ operator on both sides, we obtain:
\begin{align*}
    \pth{B(\bm{A})^T \otimes \text{Att}_1(\widehat{\bm{X}})}\text{vec}(\bm{X}) = \pth{Q_1^T \otimes \bm{I}} \text{vec}(X)
\end{align*}
Since, we want this to be true for all $X$, clearly,
\begin{align*}
     B(\bm{A})^T \otimes \text{Att}_1(\widehat{\bm{X}}) &=  Q_1^T \otimes \bm{I} \\
     \implies \text{rank}(\text{Att}_1(\widehat{\bm{X}})) &= \text{rank}(Q_1^T) \cdot \text{rank}(\bm{I}) / \text{rank}(B(\bm{A}))
\end{align*}
Note that we have used $\text{rank}(\bm{A} \otimes \bm{B}) = \text{rank}(\bm{A}) \cdot \text{rank}(\bm{B})$ We know that $\text{rank}(Q_1^T)=d_1$ and hence, since $\text{rank}(B(\bm{A}))\leq d_1$, then clearly, 
\begin{align*}
    \text{rank}(\text{Att}_1(\widehat{\bm{X}})) \geq N
\end{align*}
For this to be satisfied, we can see that the query and the key matrices both have to be of size at-least $(N+d)N$. Moreover, the value matrix weights are then going to be of size $(N+d)d_1$. Hence, the total dimensions required for this is $2(N+d)N + (N+d)d_1$. We then argue about the second case which is what happens if $d_1 > d$. For the argument, we assume that this spans two actual polynimial filters that is $d< d_1<2d$, but we will then show that the argument will hold even if $d_1$ was arbitarty. Due to this, we obtain that, for equality between the representations when the number of heads are greater than that of the degree of the polynomial filter ($H>k$),  the following: 
\begin{align*}
\begin{bmatrix}\text{Att}_1(\widehat{\bm{X})}\bm{X}B_1(\bm{A})\bm{Q}_{1} & \text{Att}_1(\widehat{\bm{X})}\bm{X}B_1(\bm{A})\bm{Q}_2
    \end{bmatrix} = \begin{bmatrix}
        \bm{X} & \bm{A}\bm{X}\bm{\widehat{Q}_{ 2}}
    \end{bmatrix}
\end{align*}
Note that $\bm{Q_1} \in \mathbb{R}^{d_1 \times d}, \bm{Q_2} \in \mathbb{R}^{d_1 \times (d_1-d)}, \bm{\widehat{Q}_{ 2}} \in \mathbb{R}^{d \times d_1}$, with ranks $d, d_1, d_1-d$ respectively. Now we just equate them,
\begin{align*}
\pth{(B_1(\bm{A})\bm{Q}_1)^T\otimes  \text{Att}_1(\widehat{\bm{X})}} \cdot \text{vec}\bm{X} &= \pth{\bm{I}_{d \times d} \otimes \bm{I}_{N\times N}} \cdot \text{vec}\bm{X} \\
\pth{(B_1(\bm{A})\bm{Q}_2)^T\otimes  \text{Att}_1(\widehat{\bm{X})}} \cdot \text{vec}\bm{X} &= \pth{\bm{\hat{Q}}_2 \otimes \bm{A}} \cdot \text{vec}\bm{X} 
\end{align*}
Now we use the same rank argument as before, to conclude that $\text{rank}(\text{Att}_1(\widehat{\bm{X})}) \geq N$, which again implies that the key and query weight matrices have weights $(N+d)N$ respectively. 

We make two generalizations now. We would first like to note that this argument also holds when $d_1 \geq (k-1)d$ and hence, the above proof works for any dimensions. We would also like to note that this while this proof is done for the first head of MHA, it also generlizes to other heads. The proof holds similarly, and constructively that is for the second head of MHA, one can make the similar argument as that of the first with the only difference being that one needs to udnerstand the dimensions of the output polynomial filter that is which dimensions of the second head of MHA corresponds to which dimensiond of the polynomial filter. Then by repeating the argument of the ranks, one can obtain that again, key and query weight matrices have weights $(N+d)N$ respectively. This argument can be repeated for the third head and so on to finally conclude that the number of dimensions needed to represent this is $Hn(N+d) + (d+N)(kd)$, when $H \geq k$. Clearly, this is minimal when $H=k$.

\paragraph{IHA.} We present a construction that achieves comparable parametric complexity while requiring only $\lceil \sqrt{k} \rceil$ heads. The proof proceeds by explicit construction. We note that $\bm{\widetilde{X}^{(1)}}$ denotes the representation after the first layer of attention. 
\begin{align*}
\bm{W}_{K,\texttt{IHA}}^{(1, h)} &= \begin{bmatrix}
        \bm{0}_{d\times N} \\
        (A^{h-1})^\top
    \end{bmatrix} ~\forall~h\in \{1, 2, \cdots, \lceil \sqrt{k} \rceil\} \\
    \bm{W}_{Q,\texttt{IHA}}^{(1, h)} &= \begin{bmatrix}
        \bm{0}_{d\times N} \\
        A^{(h-1) \cdot \lceil\sqrt{k}\rceil}
    \end{bmatrix} ~\forall~h\in \{1, 2, \cdots, \lceil \sqrt{k} \rceil\} \\
     \bm{W}_{V,\texttt{IHA}}^{(1, h)} &= \begin{bmatrix}
        \bm{L}^h_{d \times d\lceil \sqrt{k} \rceil} \\
        \bm{0}_{N \times d\lceil \sqrt{k} \rceil}
    \end{bmatrix}  \in \mathbb{R}^{(N+d)\times \lceil d\sqrt{k} \rceil } \\
\text{where, } \bm{L}^{(1, h)}_{d \times d\lceil \sqrt{k} \rceil} &= \begin{bmatrix}
    \bm{0}_{d \times (h-1)d} & I_{d\times d} &\bm{0}_{d \times  \lceil \sqrt{k} \rceil -hd}  
\end{bmatrix} \forall \ h~\in \{1,\ 2,\ \cdots,\ \lceil \sqrt{k} \rceil\} \\
&\quad\text{with the convention that } \bm{0}_{d \times 0} \text{ represents an empty vector.} \\
\text{moreover, } m_{k, h}^{(1)} &= 1 \ \forall k, h \in \{1, 2, \cdots, \ceil{\sqrt{k}} \}
\end{align*}
Hence, after one layer of attention we obtain the following. Using bracket notation $[\cdot, \cdot]$ for concatenation:
\begin{align*}
    \bm{\widetilde{X}^{(1)}} &= \Big[\widetilde{\bm{X}}^{(1,1)}_{\texttt{IHA}},\; \widetilde{\bm{X}}^{(1,2)}_{\texttt{IHA}},\; \ldots,\; \widetilde{\bm{X}}^{(1,\ceil{\sqrt{k}})}_{\texttt{IHA}}\Big]
\end{align*}
where each $\widetilde{\bm{X}}^{(1,h)}_{\texttt{IHA}}$ aggregates over all key heads.
\begin{align*}
     \bm{\widetilde{X}^{(1)}} &= \begin{bmatrix}
        \bm{X} & \bm{A}\bm{X} & \cdots & \bm{A}^{\ceil{\sqrt{k}}^2-1}\bm{X}
    \end{bmatrix}
\end{align*}

Hence, under this construction, the number of parameters is  $2n(N+d) \lceil \sqrt{k} \rceil + d(d+N)\ceil{\sqrt{k}}^2) + \ceil{\sqrt{k}}^2$

\paragraph{Example.} We present an example here to solidify the intuition of why such a construction helps. We assume that $k=4$. Hence, using the above constructions, the query, key, value matrices are defined as follows.
\begin{align*}
        \bm{W}_{Q,\texttt{IHA}}^{(1, 1)} &= \begin{bmatrix}
        \bm{0}_{d\times N} \\
        \bm{I}_{N \times N}
    \end{bmatrix}  \quad     \bm{W}_{Q,\texttt{IHA}}^{(1, 2)} = \begin{bmatrix}
        \bm{0}_{d\times N} \\
        \pth{A^2}^\top
    \end{bmatrix}  \\
    \bm{W}_{K,\texttt{IHA}}^{(1, 1)} &= \begin{bmatrix}
        \bm{0}_{d\times N} \\
        \bm{I}_{N \times N}
    \end{bmatrix}  \quad     \bm{W}_{K,\texttt{IHA}}^{(1, 2)} = \begin{bmatrix}
        \bm{0}_{d\times N} \\
        \pth{A}^\top
    \end{bmatrix}  \\
    \bm{W}_{V,\texttt{IHA}}^{(1, 1)} &= \begin{bmatrix}
        \bm{I}_{d \times d} & \bm{0}_{d\times d} \\
        \bm{0}_{N \times  d} & \bm{0}_{N \times  d}
    \end{bmatrix} \quad      \bm{W}_{V,\texttt{IHA}}^{(1, 2)} = \begin{bmatrix}
        \bm{0}_{d \times d} & \bm{I}_{d\times d} \\
        \bm{0}_{N \times  d} & \bm{0}_{N \times  d}
    \end{bmatrix} \\
    m_{k, h}^{(1)} &=1 \ \forall \ k, h \in \{1, 2\}
\end{align*}
We can see that first head of attention computes to
\begin{align*}
    Z_1 &= \begin{bmatrix}
        \pth{\widehat{\bm{X}}\bm{W}_{Q,\texttt{IHA}}^{(1, 1)} (\bm{W}_{K,\texttt{IHA}}^{(1, 1)})^\top \widehat{\bm{X}}^\top} &         \pth{\widehat{\bm{X}}\bm{W}_{Q,\texttt{IHA}}^{(1, 1)} (\bm{W}_{K,\texttt{IHA}}^{(1, 2)})^\top \widehat{\bm{X}}^\top}
    \end{bmatrix} \begin{bmatrix}
        \widehat{X}  \bm{W}_{V,\texttt{IHA}}^{(1, 1)} \\
        \widehat{X}  \bm{W}_{V,\texttt{IHA}}^{(1, 2)}
    \end{bmatrix}  \\
    &= \begin{bmatrix}
        \bm{I} & \bm{A}
    \end{bmatrix} \begin{bmatrix}
        \bm{X} & \bm{0} \\
        \bm{0} & \bm{X}
    \end{bmatrix} \\
    &= \begin{bmatrix}
        \bm{X} & \bm{A} \bm{X}
    \end{bmatrix}
\end{align*}
Similarly, for the second layer of attention, we obtain
\begin{align*}
    Z_2 &= \begin{bmatrix}
        \pth{\widehat{\bm{X}}\bm{W}_{Q,\texttt{IHA}}^{(1, 2)} (\bm{W}_{K,\texttt{IHA}}^{(1, 1)})^\top \widehat{\bm{X}}^\top} &         \pth{\widehat{\bm{X}}\bm{W}_{Q,\texttt{IHA}}^{(1, 2)} (\bm{W}_{K,\texttt{IHA}}^{(1, 2)})^\top \widehat{\bm{X}}^\top}
    \end{bmatrix} \begin{bmatrix}
        \widehat{X}  \bm{W}_{V,\texttt{IHA}}^{(1, 1)} \\
        \widehat{X}  \bm{W}_{V,\texttt{IHA}}^{(1, 2)}
    \end{bmatrix}  \\
    &= \begin{bmatrix}
        \bm{A}^2 & \bm{A}^3
    \end{bmatrix} \begin{bmatrix}
        \bm{X} & \bm{0} \\
        \bm{0} & \bm{X}
    \end{bmatrix} \\
    &= \begin{bmatrix}
        \bm{A}^2\bm{X} & \bm{A}^3 \bm{X}
    \end{bmatrix}
\end{align*}
Hence, on concatenating the the embeddings from both the heads, we obtain
\begin{align*}
     \bm{\widetilde{X}^{(1)}} &= \begin{bmatrix}
         \bm{X} & \bm{A} \bm{X} & \bm{A}^2\bm{X} & \bm{A}^3 \bm{X}
     \end{bmatrix}
\end{align*}
\end{proof}

\paragraph{IHA.}
We present a construction that requires only $H\coloneqq \lceil \sqrt{k}\rceil$ heads. We set the number of pseudo heads to be $P\coloneqq H$. Note that the input to \iha~is the same as \mha~as defined below. 
\[
\widehat{\bm{X}} = 
\begin{bmatrix}
\bm{X} & \bm{I}
\end{bmatrix},
\]
The proof proceeds by explicit construction. For every base head index $m\in\{1,2,\ldots,H\}$ define
\begin{align*}
\bm{W}_{K,\texttt{IHA}}^{(1,m)} &= 
\begin{bmatrix}
\bm{0}_{d\times N}\\
\pth{A^{m-1}}^\top
\end{bmatrix}, \\
\bm{W}_{Q,\texttt{IHA}}^{(1,m)} &=
\begin{bmatrix}
\bm{0}_{d\times N}\\
A^{(m-1)\cdot H}
\end{bmatrix}, \\
\bm{W}_{V,\texttt{IHA}}^{(1,m)} &= 
\begin{bmatrix}
\bm{L}^{(1,m)}_{d\times dH}\\
\bm{0}_{N\times dH}
\end{bmatrix}\in\mathbb{R}^{(N+d)\times dH},
\end{align*}
where $\bm L^{(1,m)}_{d\times dH}$ routes into the $m$-th $d$-block (same selector trick as before):
\begin{align*}
\bm{L}^{(1,m)}_{d\times dH}
\;\coloneqq\;
\begin{bmatrix}
\bm 0_{d\times (m-1)d} & I_{d\times d} & \bm 0_{d\times (H-m)d}
\end{bmatrix},
\qquad \forall\ m\in\{1,\ldots,H\},
\end{align*}
with the convention that $\bm 0_{d\times 0}$ is empty.

We choose pseudo-head coefficients to be one-hot routers so that, \emph{inside each head $h$}, the $P=H$ pseudo-heads instantiate the same “key heads”:
\begin{align*}
\alpha^Q_{m,h,j} &\coloneqq \mathbbm{1}[m=h]\cdot \mathbbm{1}[j=1], \\
\alpha^K_{m,h,j} &\coloneqq \mathbbm{1}[m=j], \\
\alpha^V_{m,h,j} &\coloneqq \mathbbm{1}[m=j],
\qquad
\forall\ m,h\in\{1,\ldots,H\},\ \forall j\in\{1,\ldots,P\}.
\end{align*}
Thus, for each head $h$ and pseudo-head $j$,
\begin{align*}
\widetilde{\bm Q}_{h,j}
&=\sum_{m=1}^H\alpha^Q_{m,h,j}\,\widehat{\bm X}\bm W_{Q,\texttt{IHA}}^{(1,m)}
=\mathbbm{1}[j=1]\ \widehat{\bm X}\bm W_{Q,\texttt{IHA}}^{(1,h)}, \\
\widetilde{\bm K}_{h,j}
&=\sum_{m=1}^H\alpha^K_{m,h,j}\,\widehat{\bm X}\bm W_{K,\texttt{IHA}}^{(1,m)}
=\widehat{\bm X}\bm W_{K,\texttt{IHA}}^{(1,j)}, \\
\widetilde{\bm V}_{h,j}
&=\sum_{m=1}^H\alpha^V_{m,h,j}\,\widehat{\bm X}\bm W_{V,\texttt{IHA}}^{(1,m)}
=\widehat{\bm X}\bm W_{V,\texttt{IHA}}^{(1,j)}.
\end{align*}
For each head $h$, \iha~stacks pseudo-queries and keys row-wise:
\begin{align*}
\overline{\bm Q}_{h}\coloneqq\left[\widetilde{\bm Q}_{h,1}^\top;\ldots;\widetilde{\bm Q}_{h,H}^\top\right]^\top,
\quad
\overline{\bm K}_{h}\coloneqq\left[\widetilde{\bm K}_{h,1}^\top;\ldots;\widetilde{\bm K}_{h,H}^\top\right]^\top,
\quad
\overline{\bm V}_{h}\coloneqq\left[\widetilde{\bm V}_{h,1}^\top;\ldots;\widetilde{\bm V}_{h,H}^\top\right]^\top.
\end{align*}
\iha~computes $\overline{\bm P}_{h}=\softmax(\overline{\bm Q}_{h}\overline{\bm K}_{h}^\top)\overline{\bm V}_{h}$.
Consider the output corresponding to the \emph{first pseudo query} (i.e., the first $N$ rows of $\overline{\bm P}_h$),
which we denote by $\bm P_{h,1}\in\mathbb{R}^{N\times dH}$. By construction, $\widetilde{\bm Q}_{h,1}=\widehat{\bm X}\bm W_{Q,\texttt{IHA}}^{(1,h)}$
and $\widetilde{\bm K}_{h,j}=\widehat{\bm X}\bm W_{K,\texttt{IHA}}^{(1,j)}$ for all $j$,
so the same “aggregate over all key heads” block product:
\begin{align*}
\bm P_{h,1}
&=
\begin{bmatrix}
\pth{\widehat{\bm{X}}\bm{W}_{Q,\texttt{IHA}}^{(1,h)} (\bm{W}_{K,\texttt{IHA}}^{(1,1)})^\top \widehat{\bm{X}}^\top}
& \cdots &
\pth{\widehat{\bm{X}}\bm{W}_{Q,\texttt{IHA}}^{(1,h)} (\bm{W}_{K,\texttt{IHA}}^{(1,H)})^\top \widehat{\bm{X}}^\top}
\end{bmatrix}
\begin{bmatrix}
\widehat{\bm X}\bm W_{V,\texttt{IHA}}^{(1,1)}\\
\vdots\\
\widehat{\bm X}\bm W_{V,\texttt{IHA}}^{(1,H)}
\end{bmatrix}.
\end{align*}
Using the definitions of $\bm W_Q^{(1,h)}$ and $\bm W_K^{(1,j)}$,
\begin{align*}
\widehat{\bm{X}}\bm{W}_{Q,\texttt{IHA}}^{(1,h)} (\bm{W}_{K,\texttt{IHA}}^{(1,j)})^\top \widehat{\bm{X}}^\top
\;=\;
A^{(h-1)H}\,A^{j-1}
\;=\;
A^{(h-1)H+(j-1)}.
\end{align*}
Moreover, $\widehat{\bm X}\bm W_{V,\texttt{IHA}}^{(1,j)}$ routes $\bm X$ into the $j$-th $d$-block
inside the $dH$-dimensional head space. Therefore,
\begin{align*}
\bm P_{h,1}
&=
\begin{bmatrix}
A^{(h-1)H}\bm X & A^{(h-1)H+1}\bm X & \cdots & A^{(h-1)H+(H-1)}\bm X
\end{bmatrix}.
\end{align*}
\iha~then collapses the $HP$ pseudo-outputs down to $H$ heads via $\bm R\in\mathbb{R}^{H\times HP}$.
We choose $\bm R$ to select \emph{only} pseudo $j=1$ from each head $h$:
\begin{align*}
\bm R_{h,(h-1)H+1}=1,\qquad
\bm R_{h,(h-1)H+j}=0\ \ \forall j\in\{2,\ldots,H\},\qquad
\bm R_{h,(h'-1)H+j}=0\ \ \forall h'\neq h.
\end{align*}
Hence the collapsed head output is
\begin{align*}
\bm O_h \;=\; \bm P_{h,1}
\;=\;
\begin{bmatrix}
A^{(h-1)H}\bm X & A^{(h-1)H+1}\bm X & \cdots & A^{(h-1)H+(H-1)}\bm X
\end{bmatrix}.
\end{align*}

\medskip
\noindent
\textbf{Concatenate heads.}
Finally on concatenating representations from different heads, we obtain,
\begin{align*}
\bm{\widetilde{X}^{(1)}}
&=
\Big[\bm O_1,\bm O_2,\ldots,\bm O_H\Big]
=
\begin{bmatrix}
\bm X & \bm A\bm X & \cdots & \bm A^{H^2-1}\bm X
\end{bmatrix}.
\end{align*}
Thus the construction realizes all powers up to $A^{H^2-1}$ in one layer.
If $H^2>k$, the extra $(H^2-k)$ blocks may be treated as padding (or zeroed by an output mask).

Hence, under this construction, the number of parameters is  $2n(N+d) \lceil \sqrt{k} \rceil + d(d+N)\ceil{\sqrt{k}}^2) + 4\ceil{\sqrt{k}}^3$

\paragraph{Example ($k=4$).}
Let $k=4$, so $H=P=2$.
Then
\[
\bm W_{Q}^{(1,1)}=\begin{bmatrix}0\\I\end{bmatrix},\quad
\bm W_{Q}^{(1,2)}=\begin{bmatrix}0\\A^{2}\end{bmatrix},\quad
\bm W_{K}^{(1,1)}=\begin{bmatrix}0\\I\end{bmatrix},\quad
\bm W_{K}^{(1,2)}=\begin{bmatrix}0\\A^\top\end{bmatrix},
\]
and $\bm W_V^{(1,1)},\bm W_V^{(1,2)}$ route into the first/second $d$-block, exactly as in the old proof.
Choose $\alpha$ one-hot as above, and choose $\bm R$ to pick pseudo $j=1$ from each head.
Then head $h=1$ outputs $[\bm X,\bm A\bm X]$, head $h=2$ outputs $[\bm A^2\bm X,\bm A^3\bm X]$,
and concatenation yields
\begin{align*}
\bm{\widetilde X^{(1)}}=
\begin{bmatrix}
\bm X & \bm A\bm X & \bm A^2\bm X & \bm A^3\bm X
\end{bmatrix}.
\end{align*}

\subsection{Representing Count Permutation Match 3 (CPM-3)}
\label{app:cpm3}

\begin{theorembox}[label=thm:perutationmatch3_proof]
{(Count Permutation Match-3): }  Let $\nmax$ denote the maximum number of tokens that can be processed by the model in the worst case. There exists a one-layer transformer with \emph{interleaved-head attention} (\iha) that can represent the permutation match-3 task using $\ceil{\sqrt{\nmax}}$ attention heads. The number of parameters required by this IHA construction is upper bounded by $37 \nmax^2 \sqrt{\nmax} + \nmax^2 (\nmax - 1) + \nmax^2$. In contrast, the best currently known construction based on \emph{multi-head attention} (\mha) requires $\nmax$ attention heads, and its parameter count is lower bounded by $3 \nmax^3 + \nmax^2 (\nmax - 1) + \nmax^2$. Throughout, we assume the vocabulary size is at most on the order of the maximum sequence length, i.e., $|\mathcal{V}|=O(\nmax)$.
\end{theorembox}

\paragraph{Count Permutation Match-3 (CPM-3):} We define a task denoted as Count Permutation Match-3 (CPM-3) where the goal is given a sequence of natural numbers denoted as $\{x_i \}_{i \in \mathbb{N}}$, the goal is to be able to count the number of occurrences of a specific function, defined as follows. Let us define the number of triples ($i, j_1, j_2)$ where $i$ denotes the token in consideration, that satisfy:
\begin{align*}
\mathrm{CPM}_i(3)
    &= \mathrm{Count}\Bigl( \forall j_1, j_2 \ :
        \phi(x_i, x_{j_1}, x_{j_2})=0 \Bigr),
\end{align*}
where,
\begin{align*}
\phi(x_i, x_{j_1}, x_{j_2})
    &:= x_i + G x_{j_1} + x_{j_2}
     \;\mathrm{mod}\; M \\
     &\quad \text{ where } M \text{ is an arbitrary number and } G \text{ is another number such that} G>2M.
\end{align*}

The above condition is required to make sure that the function is not permutation invariant. That is, if $x_{j_1} \neq x_{j_2}$ then, one can clearly see that:
\begin{align*}
    \phi(x_i, x_{j_1}, x_{j_2}) \neq \phi(x_i, x_{j_2}, x_{j_1})
\end{align*}

\medskip
\textbf{Example.}  
Let the sequence be $(1,2,3)$ and let $G=10$.  
Then we can compute $x_{j_1} + x_{j_2}$ as follows:
\begin{align*}
11,\; 12,\; 13,\; 21,\; 22,\; 23,\; 31,\; 32,\; 33
\end{align*}

Then we can finally compute $x_i + x_{j_1} + x_{j_2}$ which is computed as follows:
\begin{align*}
1+ x_{j_1} + x_{j_2} &:= 12,\; 13,\; 14,\; 22,\; 23,\; 24,\; 32,\; 33,\; 34 \\ 
2+ x_{j_1} + x_{j_2} &:= 13,\; 14,\; 15,\; 23,\; 24,\; 25,\; 33,\; 34,\; 35 \\
3+ x_{j_1} + x_{j_2} &:= 14,\; 15,\; 16,\; 24,\; 25,\; 26,\; 34,\; 35,\; 36 
\end{align*}
\begin{align*}
\mathrm{CPM}_1(3) &:= 3 \\ 
\mathrm{CPM}_2(3) &:= 3 \\ 
\mathrm{CPM}_3(3) &:= 3
\end{align*}

\begin{proof} The proof will follow by construction as before. That is, we show that a representative solution exists, which is constructed via different components such as attention, MLP etc. We first describe the Encoder and Positional Encodings below

\paragraph{IHA.} We first show the \textbf{IHA} construction and then proceed to the \textbf{MHA} construction.

We prove realizability by explicit construction. The construction has three components:
(i) an encoder/positional encoding map, (ii) one \iha~attention layer (with pseudo-heads) that brings all
symbols into a single vector space at each position via structured cyclic shifts, and (iii) an MLP that computes
$\mathrm{CPM}_i(3)$ by enumerating ordered pairs $(j_1,j_2)$ and aggregating indicators of the constraint
$\phi(x_i,x_{j_1},x_{j_2})\equiv 0\ (\mathrm{mod}\ M)$.

\paragraph{Encoder and Positional Encoding.}
The input is a length-$N_{\max}$ sequence of natural numbers $\{x_i\}_{i=1}^{N_{\max}}$.
Assume the encoder and positional encoding produce
\begin{align*}
\hat{\bm X}=[\,\bm X,\ \bm I\,],
\end{align*}
where $\bm X\in\mathbb{R}^{N_{\max}\times 1}$ stores the scalar token values, and
$\bm I\in\mathbb{R}^{N_{\max}\times N_{\max}}$ is the identity positional encoding. Hence
\begin{align*}
\hat{\bm X}\in\mathbb{R}^{N_{\max}\times (N_{\max}+1)}.
\end{align*}

\paragraph{IHA Attention (Layer 1).}
Let $\bm P\in\mathbb{R}^{N_{\max}\times N_{\max}}$ denote the cyclic permutation matrix and
$\bm e_1=[1,\bm 0_{1\times N_{\max}}]^\top$.
Set
\[
H\coloneqq \ceil{\sqrt{N_{\max}}},\qquad P\coloneqq H.
\]
We construct a single new-IHA layer with $H$ heads and $P$ pseudos per head. For each index $m\in\{1,\dots,H\}$ define
\begin{align*}
\bm W_Q^{(m)} &\coloneqq 
\begin{bmatrix}
\bm 0_{1\times N_{\max}}\\
\bm P^{(m-1)H}
\end{bmatrix},
\qquad
\bm W_K^{(m)} \coloneqq 
\begin{bmatrix}
\bm 0_{1\times N_{\max}}\\
\big(\bm P^{m-1}\big)^\top
\end{bmatrix},
\end{align*}
and define value projections that route the scalar symbol coordinate into the $m$-th coordinate of an $H$-dimensional
value space:
\begin{align*}
\bm{W}_{V}^{(1)} =
\begin{bmatrix}
\bm{e_1} \cdot {\ceil{\sqrt{\nmax}}} & \bm{0}_{ (\nmax+1) \times (\ceil{\sqrt{\nmax}}-1)}
\end{bmatrix},
\end{align*}
\begin{align*}
\bm{W}_{V,\texttt{IHA}}^{(1, \ceil{\sqrt{\nmax}})} =
\begin{bmatrix}
\bm{0}_{ (\nmax+1) \times (\ceil{\sqrt{\nmax}}-1)} & \bm{e_1} \cdot {\ceil{\sqrt{\nmax}}} 
\end{bmatrix},
\end{align*}
\begin{align*}
\bm{W}_{V,\texttt{IHA}}^{(1, h)} =
\begin{bmatrix}
\bm{0}_{ (\nmax+1) \times (h-1)} & \bm{e_1} \cdot {\ceil{\sqrt{\nmax}}} & \bm{0}_{ (\nmax+1) \times (\ceil{\sqrt{\nmax}} -h)}
\end{bmatrix},
\end{align*}

\medskip
\noindent
\textbf{Pseudo-head mixing.}
Let $\alpha^Q,\alpha^K,\alpha^V\in\mathbb{R}^{H\times H\times P}$ be the new-IHA mixing coefficients.
We set them to one-hot routers:
\begin{align*}
\alpha^Q_{m,h,j} &\coloneqq \mathbbm{1}[m=h]\cdot \mathbbm{1}[j=1],\\
\alpha^K_{m,h,j} &\coloneqq \mathbbm{1}[m=j],\\
\alpha^V_{m,h,j} &\coloneqq \mathbbm{1}[m=j],
\qquad
\forall m,h\in\{1,\dots,H\},\ \forall j\in\{1,\dots,P\}.
\end{align*}
Consequently, for each head $h$ and pseudo $j$,
\begin{align*}
\widetilde{\bm Q}_{h,j}
&=\sum_{m=1}^H\alpha^Q_{m,h,j}\ \hat{\bm X}\bm W_Q^{(m)}
=\mathbbm{1}[j=1]\ \hat{\bm X}\bm W_Q^{(h)},\\
\widetilde{\bm K}_{h,j}
&=\sum_{m=1}^H\alpha^K_{m,h,j}\ \hat{\bm X}\bm W_K^{(m)}
=\hat{\bm X}\bm W_K^{(j)},\\
\widetilde{\bm V}_{h,j}
&=\sum_{m=1}^H\alpha^V_{m,h,j}\ \hat{\bm X}\bm W_V^{(m)}
=\hat{\bm X}\bm W_V^{(j)}.
\end{align*}

\medskip
\noindent
\textbf{Pseudo-major stacking and attention.}
For each $h\in\{1,\dots,H\}$ define the stacked pseudo matrices
\begin{align*}
\overline{\bm Q}_h \coloneqq \left[\widetilde{\bm Q}_{h,1}^\top;\ldots;\widetilde{\bm Q}_{h,P}^\top\right]^\top,\quad
\overline{\bm K}_h \coloneqq \left[\widetilde{\bm K}_{h,1}^\top;\ldots;\widetilde{\bm K}_{h,P}^\top\right]^\top,\quad
\overline{\bm V}_h \coloneqq \left[\widetilde{\bm V}_{h,1}^\top;\ldots;\widetilde{\bm V}_{h,P}^\top\right]^\top.
\end{align*}
Let
\[
\bm S_h\coloneqq \frac{1}{\sqrt{H}}\ \overline{\bm Q}_h\overline{\bm K}_h^\top,\qquad
\overline{\bm P}_h \coloneqq \softmax(\bm S_h)\ \overline{\bm V}_h.
\]
We set the softmax temperature to $0$ (hard attention), so that the attention map implements the unique routing
induced by the permutation structure. Define $\bm P_{h,1}\in\mathbb{R}^{N_{\max}\times H}$ to be the output block of $\overline{\bm P}_h$ corresponding to
the first pseudo (i.e., the rows aligned with $\widetilde{\bm Q}_{h,1}$). Under hard attention, the construction
ensures that $\widetilde{\bm Q}_{h,1}$ routes to each pseudo-key $\widetilde{\bm K}_{h,j}$ according to the cyclic
shifts, yielding
\begin{align*}
\bm P_{h,1}
&=
\sum_{j=1}^{H}\bm P^{(h-1)H+(j-1)}\ \hat{\bm X}\bm W_V^{(j)}.
\end{align*}
Since $\hat{\bm X}\bm W_V^{(j)}$ places the scalar symbol value into the $j$-th coordinate of $\mathbb{R}^{H}$,
it follows that $\bm P_{h,1}$ contains, in an $H$-dimensional value space, the $H$ cyclic shifts
\begin{align*}
\begin{bmatrix}
    \bm P^{(h-1)H}\bm X, & \bm P^{(h-1)H+1}\bm X, & \ldots, & \bm P^{(h-1)H+(H-1)}\bm X
\end{bmatrix}    
\end{align*}
Let $\bm R\in\mathbb{R}^{H\times HP}$ be the collapse matrix in the new-IHA definition.
Choose $\bm R$ to select only pseudo $j=1$ from each head:
\begin{align*}
\bm R_{h,(h-1)H+1}=1,\qquad
\bm R_{h,(h-1)H+j}=0\ \forall j\in\{2,\dots,H\},\qquad
\bm R_{h,(h'-1)H+j}=0\ \forall h'\neq h.
\end{align*}
Then the per-head output is $\bm O_h=\bm P_{h,1}$.

\textbf{Representation after Layer 1.}
Let $\hat{\bm X}^{(1)}$ denote the representation after the new-IHA layer, obtained by concatenating heads:
\[
\hat{\bm X}^{(1)}\coloneqq [\,\bm O_1,\bm O_2,\ldots,\bm O_H\,]\in\mathbb{R}^{N_{\max}\times H^2}.
\]
By the expression for $\bm O_h=\bm P_{h,1}$, $\hat{\bm X}^{(1)}$ contains \emph{all} cyclic shifts
$\{\bm P^t\bm X\}_{t=0}^{H^2-1}$ arranged in a fixed, known indexing across its $H^2$ coordinates.
In particular, for every token position $i$, the vector $\hat{\bm X}^{(1)}[i,:]$ provides access to the entire
multiset of symbols $\{x_1,\dots,x_{N_{\max}}\}$ (with their cyclic order), within a single vector space.

\paragraph{MLP.}
Note that $\hat{\bm X}^{(1)}$ allows each token to access all $\nmax$ symbols within a single (known) coordinate system.
Hence, the \textbf{MLP} can be constructed as follows. The first linear layer is chosen to enumerate all ordered pairs
$(j_1,j_2)$ (i.e., all ${}^{\nmax}P_{2}=\nmax(\nmax-1)$ permutations, or $\nmax^2$ pairs if allowing $j_1=j_2$), and for
each such pair it forms the quantity
\[
x_i + Gx_{j_1} + x_{j_2}.
\]
This can be implemented by a linear map whose hidden width scales with the number of pairs, with weights that (i) select
$x_{j_1}$ and $x_{j_2}$ from $\hat{\bm X}^{(1)}[i,:]$ and (ii) multiply the selected $x_{j_1}$ by $G$ while leaving
$x_{j_2}$ unscaled. Furthermore, if the activation after this first layer is defined as
$f(\cdot)=\mathrm{ReLU}\bigl(1-\phi(\cdot)\bigr)$ where $\phi(\cdot)$ is the modulo-$M$ operation, then the activation
produces a nonzero output if and only if
\[
x_i + Gx_{j_1} + x_{j_2}\equiv 0\quad (\mathrm{mod}\ M).
\]
Finally, the second linear layer is responsible solely for aggregating (summing) these indicators across all ordered
pairs $(j_1,j_2)$, thereby producing $\mathrm{CPM}_i(3)$ at each token position $i$.

\paragraph{Total Parameter Count.}
Considering all parameters described above, the total parameter count (for the new-IHA construction) denoted by
$T_{\iha}$ can be upper bounded by the sum of: (i) base query/key/value projections, (ii) pseudo mixing coefficients,
(iii) the collapse matrix $\bm R$, and (iv) the MLP parameters. Concretely, with
$H=\ceil{\sqrt{\nmax}}$ and $P=H$, the attention-layer parameters satisfy
\begin{align*}
(\text{Query}+\text{Key})
&= 2\cdot H\cdot (\nmax+1)\nmax,\\
\text{Value}
&= H\cdot (\nmax+1)\cdot H = (\nmax+1)H^2,\\
\text{Pseudo Mix Coeffs}
&= 3\cdot H\cdot H\cdot P = 3H^3,\\
\text{Collapse }(\bm R)
&= H\cdot (HP) = H^3.
\end{align*}
For the MLP, the first layer requires width proportional to the number of ordered pairs, i.e.\ ${}^{\nmax}P_2$,
and thus contributes on the order of $H^2\cdot \nmax(\nmax-1)$ parameters (up to constant factors depending on the exact
hidden width choice), while the second layer aggregates these counts and contributes at most $\nmax^2$ parameters.
Putting these together,
\begin{align*}
T_{\iha}
&\;=\; (\text{Query}+\text{Key}) + \text{Value} + \text{MLP} + \text{Pseudo Mix Coeffs} + \text{Collapse} \\
&\;\le\; 2(\nmax^2+\nmax)H \;+\; (\nmax+1)H^2 \;+\; H^2\nmax(\nmax-1) \;+\; \nmax^2 \;+\; 4H^3. \\
&\;\le\;2(\nmax^2+\nmax)\ceil{\sqrt{\nmax}} \;+\; (\nmax+1)\ceil{\sqrt{\nmax}}^2 \;+\; \ceil{\sqrt{\nmax}}^2\nmax(\nmax-1) \;+\; \nmax^2 \;+\; 4\ceil{\sqrt{\nmax}}^3. \\
&\;\le\;37\nmax^{2.5} + \nmax^2(\nmax-1) + \nmax^2
\end{align*}
Substituting $H=\ceil{\sqrt{\nmax}}$ yields a polynomial bound in $\nmax$ (with the same dominant scaling coming from
the MLP term), completing the construction.

\paragraph{Example $N_{\max}=4$).}
We solidify the intuition behind the construction with a concrete instance.
Let the input tokens be $1,2,3,4$ and $\nmax=4$.
Then the encoder with positional encoding produce
\begin{align*}
\hat{\bm{X}}=
\begin{bmatrix}
1 & 1 & 0 & 0 & 0\\
2 & 0 & 1 & 0 & 0\\
3 & 0 & 0 & 1 & 0\\
4 & 0 & 0 & 0 & 1
\end{bmatrix}\in\mathbb{R}^{4\times 5}.
\end{align*}
Set $H\coloneqq \ceil{\sqrt{\nmax}}=2$ and $P\coloneqq H=2$.
Let $\bm P\in\mathbb{R}^{4\times 4}$ be the cyclic permutation matrix (shifting down by one):
\[
\bm P=
\begin{bmatrix}
0&0&0&1\\
1&0&0&0\\
0&1&0&0\\
0&0&1&0
\end{bmatrix},
\qquad
\bm e_1=[1,0,0,0,0]^\top\in\mathbb{R}^{5}.
\]
Let $\bm u_1=[1,0]^\top$ and $\bm u_2=[0,1]^\top$ denote the standard basis of $\mathbb{R}^{2}$.

\medskip
\noindent
\textbf{Query, Keys and Values.}
We define the query and key matrices for $m\in\{1,2\}$:
\begin{align*}
\bm{W}_{Q,\texttt{IHA}}^{(1,m)}
&=
\begin{bmatrix}
\bm 0_{1\times \nmax}\\
\bm P^{(m-1)H}
\end{bmatrix},
&
\bm{W}_{K}^{(1,m)}
&=
\begin{bmatrix}
\bm 0_{1\times \nmax}\\
\big(\bm P^{m-1}\big)^\top
\end{bmatrix},
\end{align*}
Hence, explicitly,
\begin{align*}
\bm{W}_{Q}^{(1,1)}=
\begin{bmatrix}
\bm 0\\
\bm I
\end{bmatrix},
\qquad
\bm{W}_{Q}^{(1,2)}=
\begin{bmatrix}
\bm 0\\
\bm P^{2}
\end{bmatrix},
\end{align*}
\begin{align*}
\bm{W}_{K}^{(1,1)}=
\begin{bmatrix}
\bm 0\\
\bm I
\end{bmatrix},
\qquad
\bm{W}_{K}^{(1,2)}=
\begin{bmatrix}
\bm 0\\
\bm P^{\top}
\end{bmatrix},
\end{align*}
and
\begin{align*}
\bm{W}_{V}^{(1,1)}=2\,\bm e_1\,\bm u_1^\top
=
\begin{bmatrix}
2 & 0\\
0&0\\
0&0\\
0&0\\
0&0
\end{bmatrix},
\qquad
\bm{W}_{V,\texttt{IHA}}^{(1,2)}=2\,\bm e_1\,\bm u_2^\top
=
\begin{bmatrix}
0 & 2\\
0&0\\
0&0\\
0&0\\
0&0
\end{bmatrix}.
\end{align*}

\medskip
\noindent
\textbf{Pseudo-head mixing.}
In \iha, each head $h\in\{1,2\}$ forms pseudos $j\in\{1,2\}$ via $\alpha^Q,\alpha^K,\alpha^V\in\mathbb{R}^{H\times H\times P}$.
We choose one-hot mixing:
\begin{align*}
\alpha^Q_{m,h,j} &\coloneqq \mathbbm{1}[m=h]\mathbbm{1}[j=1],&
\alpha^K_{m,h,j} &\coloneqq \mathbbm{1}[m=j],&
\alpha^V_{m,h,j} &\coloneqq \mathbbm{1}[m=j].
\end{align*}
Therefore, for each head $h$,
\begin{align*}
\widetilde{\bm Q}_{h,1}=\hat{\bm X}\bm W_Q^{(1,h)},\quad \widetilde{\bm Q}_{h,2}=\bm 0,
\qquad
\widetilde{\bm K}_{h,1}=\hat{\bm X}\bm W_K^{(1,1)},\quad \widetilde{\bm K}_{h,2}=\hat{\bm X}\bm W_K^{(1,2)},
\\
\widetilde{\bm V}_{h,1}=\hat{\bm X}\bm W_V^{(1,1)},\quad \widetilde{\bm V}_{h,2}=\hat{\bm X}\bm W_V^{(1,2)}.
\end{align*}

\medskip
\noindent
\textbf{Stacking and attention (per head).}
New-IHA stacks pseudos row-wise:
\[
\overline{\bm Q}_h=\begin{bmatrix}\widetilde{\bm Q}_{h,1}\\ \widetilde{\bm Q}_{h,2}\end{bmatrix},\quad
\overline{\bm K}_h=\begin{bmatrix}\widetilde{\bm K}_{h,1}\\ \widetilde{\bm K}_{h,2}\end{bmatrix},\quad
\overline{\bm V}_h=\begin{bmatrix}\widetilde{\bm V}_{h,1}\\ \widetilde{\bm V}_{h,2}\end{bmatrix}.
\]
Let $\overline{\bm P}_h=\softmax(\overline{\bm Q}_h\overline{\bm K}_h^\top)\overline{\bm V}_h$.
If the softmax temperature is set to $0$, the attention reduces to hard attention induced by the permutation structure.

\medskip
\noindent
\textbf{Head $h=1$.}
Since $\widetilde{\bm Q}_{1,1}=\hat{\bm X}\bm W_Q^{(1,1)}$ and the two pseudo-keys correspond to $\bm I$ and $\bm P$,
the first-pseudo output of head $1$ (denoted $\bm P_{1,1}$) takes the form
\begin{align*}
\bm P_{1,1}
&=
2\,\bm I\cdot\big(\hat{\bm X}\bm W_V^{(1,1)}\big)
\;+\;
2\,\bm P\cdot\big(\hat{\bm X}\bm W_V^{(1,2)}\big).
\end{align*}
Compute the value projections:
\begin{align*}
\hat{\bm X}\bm W_V^{(1,1)}
&=
\begin{bmatrix}
2&0\\
4&0\\
6&0\\
8&0
\end{bmatrix}
\qquad
\hat{\bm X}\bm W_V^{(1,2)}
=
\begin{bmatrix}
0&2\\
0&4\\
0&6\\
0&8
\end{bmatrix}.
\end{align*}
Thus,
\begin{align*}
\bm P_{1,1}
&=
\frac{1}{2}\bm I\cdot
\begin{bmatrix}
2&0\\
4&0\\
6&0\\
8`&0
\end{bmatrix}
+
\frac{1}{2}\bm P\cdot
\begin{bmatrix}
0&\frac{1}{2}\\
0&1\\
0&\frac{3}{2}\\
0&2
\end{bmatrix}
=
\begin{bmatrix}
1&2\\
2&3\\
3&4\\
4&1
\end{bmatrix}.
\end{align*}

\medskip
\noindent
\textbf{Head $h=2$.}
Here $\widetilde{\bm Q}_{2,1}=\hat{\bm X}\bm W_Q^{(1,2)}$, so the induced shifts are $\bm P^2$ and $\bm P^3$.
Hence
\begin{align*}
\bm P_{2,1}
&=
\frac{1}{2}\,\bm P^2\cdot\big(\hat{\bm X}\bm W_V^{(1,1)}\big)
\;+\;
\frac{1}{2}\,\bm P^3\cdot\big(\hat{\bm X}\bm W_V^{(1,2)}\big)
=
\begin{bmatrix}
3&4\\
4&1\\
1&2\\
2&3
\end{bmatrix}.
\end{align*}

\medskip
\noindent
\textbf{Collapse $HP\to H$ (select pseudo $j=1$).}
New-IHA uses $\bm R\in\mathbb{R}^{H\times HP}$. Choose $\bm R$ to pick only pseudo $j=1$ from each head:
\[
\bm R_{1,1}=1,\quad \bm R_{2,3}=1,\quad \text{and all other entries are }0,
\]
so that $\bm O_1=\bm P_{1,1}$ and $\bm O_2=\bm P_{2,1}$.

\medskip
\noindent
\textbf{Concatenate heads.}
The representation after the attention layer is
\begin{align*}
\hat{\bm X}^{(1)}=[\bm O_1,\bm O_2]
=
\begin{bmatrix}
1 & 2 & 3 & 4\\
2 & 3 & 4 & 1\\
3 & 4 & 1 & 2\\
4 & 1 & 2 & 3
\end{bmatrix}.
\end{align*}
Thus each token position now contains all symbols in a fixed, known order (a cyclic listing),
which allows the subsequent MLP to enumerate ordered pairs $(j_1,j_2)$, form
$x_i + Gx_{j_1} + x_{j_2}$, apply the mod-$M$ test, and aggregate counts to compute $\mathrm{CPM}_i(3)$.

\paragraph{MHA.} We now proceed with the \textbf{MHA} construction.

\paragraph{Encoder and Positional Encoding:} The input to the network is a sequence of tokens where each token is a natural number. Concretely its defined as: $\{x_i\}_{i \in \mathbb{N}} \in \mathbb{N}$. We assume the encoder and the positional encoding to be defined such that they output the following: 
\begin{align*}
\hat{\bm{X}} = [\, \bm{X},\; \bm{I} \,],
\end{align*}
where $I$ is the identity matrix used for positional encoding, and  $X$ integer value of of the input symbol.
\begin{align*}
\bm{X} \in \mathbb{R}^{N_{\mathrm{max}} \times 1},
\qquad
\bm{I} \in \mathbb{R}^{N_{\mathrm{max}} \times N_{\mathrm{max}}},
\qquad
\hat{\bm{X}} \in \mathbb{R}^{N_{\mathrm{max}} \times (N_{\mathrm{max}}+1)}.
\end{align*}

\paragraph{Layer 1. (Attention)} We provide the sketch of what the intended outcome of the first layer is via an example and then proceed with the construction.

\textbf{Main Idea.} Similar to the prior construction, the goal of the first layer of attention is to able to permute all input symbols to obtain all the symbols in a sequence within a single vector space. For example, given an input sequence $1 \quad 2 \quad 3$, we obtain the following:
\begin{align*}
\text{Token 1:} \quad &1 \quad 2 \quad 3 \\
\text{Token 2:} \quad &2 \quad 3 \quad 2 \\
\text{Token 3:}  \quad &3 \quad 1 \quad 2.
\end{align*}

This allows us to obtain all tokens that can then be used by the MLP to obtain all the permutations needed to solve the \textbf{CPM-3} task. We first define the construction (the query, key and value matrices) in generality below and then describe an example below to make things concrete. Note that in the equation below,  $\bm{P} \in \mathbb{R}^{\nmax \times \nmax}$ denotes a cyclic permutation matrix and $\bm{e}_1 = [1, \bm{0}_{1 \times \nmax}]^T$. We first define the query matrices below. 
\begin{align*}
\bm{W}_{Q,\texttt{MHA}}^{(1, h)} =
\begin{bmatrix}
\bm{0}_{1 \times \nmax} \\
\bm{P}^{(h-1) }
\end{bmatrix}
\end{align*}
where, $h \in \{1, 2, \cdots, \nmax \}$. We now define the key matrices below.
\begin{align*}
\bm{W}_{K,\texttt{MHA}}^{(1, h)} =
\begin{bmatrix}
\bm{0}_{1 \times \nmax} \\
\bm{I}_{\nmax \times \nmax}
\end{bmatrix},
\end{align*}
where, $h \in \{1, 2, \cdots, \nmax \}$. We now define the value matrices below.
\begin{align*}
\bm{W}_{V,\texttt{IHA}}^{(1, h)} =
\begin{bmatrix}
\bm{e_1}
\end{bmatrix},
\end{align*}

where, $h \in \{1, 2, 3, \cdots, \nmax \}$. 

Hence, basis this construction, if the temperature of softmax is set to $0$ (leading to hard attention), one can obtain the following (Note that $\cdot || \cdot$ denotes the concatenation operation), and $\hat{X}^{(1)}$ denote the representation after the first layer of \iha, 

\begin{align*}
\hat{X}^{(1)} &= \bm{\hat{X}}\bm{W}_{V,\texttt{IHA}}^{(1, 1)} \ || \ \bm{P}\bm{\hat{X}}\bm{W}_{V,\texttt{IHA}}^{(1, 2)} || \ \cdots \ || \ \bm{P}^{(\nmax -1)}\bm{\hat{X}}\bm{W}_{V,\texttt{IHA}}^{(1, \nmax)}
\end{align*}

\paragraph{MLP.} Note that $\hat{X}^{(1)}$ allows for having all the $\nmax$ tokens in the same vector space. Then the \textbf{MLP} can be constructed as follows. The first layer can contain $\ceil{\nmax}^2 \times (\nmax)(\nmax-1)$ parameters where the first layer via the MLP is responsible for all ${}^{\nmax}P_{2}$ permutations along with multiplications with the right token with $G$. Furthermore, if the activation function after the first layer is defined as $f(\cdot) = \text{ReLU}(1 - \phi(\cdot))$  where $\phi(\cdot)$ the modulo-$M$ operation, then the second linear layer is responsible solely for aggregating all the resulting counts. This aggregation can be implemented by a linear layer whose parameter matrix has dimension $\max^2$.

\paragraph{Total Parameter Count.} Considering all the parameters described above, the total parameter count denoted by $T_{\mha}$ is as follows:
\begin{align*}
    T_{\mha} &= (\text{Query Parameters} + \text{Key Parameters}) + \text{Value Parameters} + \text{MLP Parameters} \\
    &= 2 \nmax^3 + \nmax ^3 + (\nmax+1)(\nmax)(\nmax-1) + \nmax^2 \\
    & > 3 \nmax^3 + (\nmax)^2(\nmax-1) + \nmax^2
\end{align*}

\end{proof}

\section{Compute and FLOP Matching for IHA}
\label{app:compute}

\paragraph{Global complexity.}
Let $N$ be the sequence length, $d$ the per-head dimension, $H$ the number of heads, and $P$ the number of pseudo-heads per head. Interleaving increases the effective sequence length from $N$ to $NP$. A global IHA layer therefore has per-head complexity
\begin{align*}
\mathcal{O}\left((NP)^2 d\right) &= \mathcal{O}(P^2N^2d),
\end{align*}
which is a factor-$P^2$ increase relative to global MHA. Accordingly, we FLOP-match all comparisons so that any improvements cannot be attributed to additional compute.

\paragraph{Hybrid local-global schedule.}
We use a local-global schedule: four layers apply sliding-window IHA with window size
\begin{align*}
W \coloneqq \frac{N}{2P^2},
\end{align*}
followed by one global-attention layer (a 4:1 ratio). In a sliding-window IHA layer, attention is computed over $NP$ query virtual tokens and $WP$ key virtual tokens, yielding per-layer cost
\begin{align*}
\mathcal{O}\left(H \cdot (NP)\cdot (WP)\cdot d\right)
&= \mathcal{O}\left(H \cdot \frac{N^2 d}{2}\right).
\end{align*}
Averaging four local layers with one global layer gives
\begin{align*}
\frac{4\cdot \mathcal{O}(HN^2d/2)+\mathcal{O}(HN^2d)}{5}
&= \frac{3}{5}\mathcal{O}(HN^2d)
\approx \mathcal{O}(HN^2d),
\end{align*}
which matches the global-attention baseline up to constant factors.

\section{Synthetic Reasoning Tasks}
\label{app:synthetic_reasoning}

In this section, we investigate whether different attention mechanisms, such as standard multi-head attention (MHA), interleaved head attention (IHA), and simplicial attention~\cite{roy2025fastsimplex}, provide measurable improvements in \emph{compositional} and \emph{multi-hop} reasoning. Following the methodology of~\cite{kozachinskiy2025strassen}, we isolate the contribution of the attention architecture using fully synthetic benchmarks with (i) ground-truth labels defined exactly by relational composition, (ii) a controlled input distribution, and (iii) dependencies that require aggregating evidence across multiple (and potentially distant) sequence positions. We describe the tasks next.

\subsection{Data and Task Formulation}
We evaluate attention mechanisms on two synthetic multi-hop reasoning tasks derived from boolean matrix composition. Both tasks share a common input format: a random $m \times m$ boolean matrix $R$, flattened into a sequence of $m^2$ binary tokens (each 0 or 1). The model must predict, for every entry $(i,j)$, whether the corresponding entry in the composed relation equals 1. This is a sequence-to-sequence binary classification problem, the input and output sequences are aligned position-wise, and training minimizes binary cross-entropy averaged over all valid positions.

\paragraph{Binary Relation Composition (2-hop).}
Given $R$, the target is $R \circ R$, where
\begin{align*}
(R \circ R)_{ij} = 1 \;\;\text{iff}\;\; \exists\, k \;\; \text{s.t.}\;\; R_{ik} = 1 \;\land\; R_{kj} = 1.
\end{align*}
This task asks whether entities $i$ and $j$ are connected by a directed path of length exactly 2 through the relation $R$. The matrix size $m$ is sampled uniformly from $\{6,\ldots,10\}$, yielding input sequences of length $m^2 \in \{36,\ldots,100\}$ that vary across examples. Each entry of $R$ is drawn i.i.d.\ from $\text{Bernoulli}(P)$ with $P = 0.325$, a value chosen empirically to yield approximately balanced positive and negative labels in $R \circ R$.

\paragraph{Ternary Relation Composition (3-hop).}
Given $R$, the target is $R \circ R \circ R$, where
\begin{align*}
(R \circ R \circ R)_{ij} = 1 \;\;\text{iff}\;\; \exists\, k,l \;\; \text{s.t.}\;\; R_{ik} = 1 \;\land\; R_{kl} = 1 \;\land\; R_{lj} = 1.
\end{align*}

This task asks whether entities $i$ and $j$ are connected by a directed path of length exactly 3 through the relation $R$. The matrix size $m$ is sampled uniformly from $\{5,\ldots,8\}$, yielding input sequences of length $m^2 \in \{25,\ldots,64\}$ that vary across examples. Each entry of $R$ is drawn i.i.d.\ from $\text{Bernoulli}(P)$ with $P = 0.264$: at higher values the 3-hop composition quickly saturates (outputs become nearly all ones), so we reduce both $m$ and $P$ to maintain approximately balanced positive and negative labels.

\subsection{Dataset Construction}
Each task uses 40{,}000 training examples, 5{,}000 validation examples, and 5{,}000 test examples. Since $m$ varies across examples, sequence lengths within a split are non-uniform; sequences within a minibatch are padded on the right to the maximum length in that batch.

\subsection{Hyperparameter Sweep and Training Protocol}

\begin{figure*}[t]
  \centering
  \begin{minipage}[t]{0.49\textwidth}
    \centering
    \includegraphics[width=\linewidth]{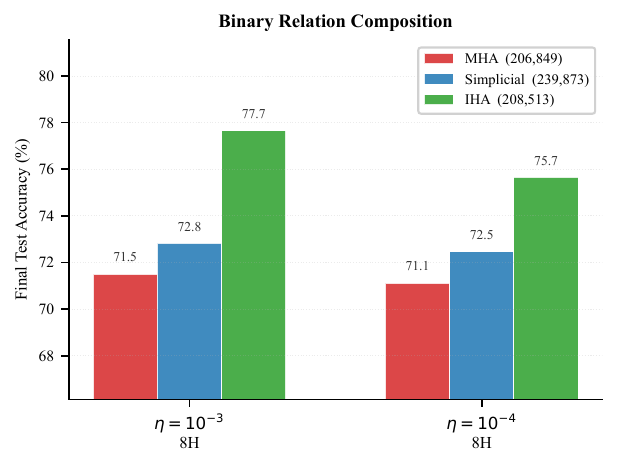}
    \caption*{(a) Binary composition: final test accuracy across learning rates.}
    \label{fig:binary-summary}
  \end{minipage}\hfill
  \begin{minipage}[t]{0.49\textwidth}
    \centering
    \includegraphics[width=\linewidth]{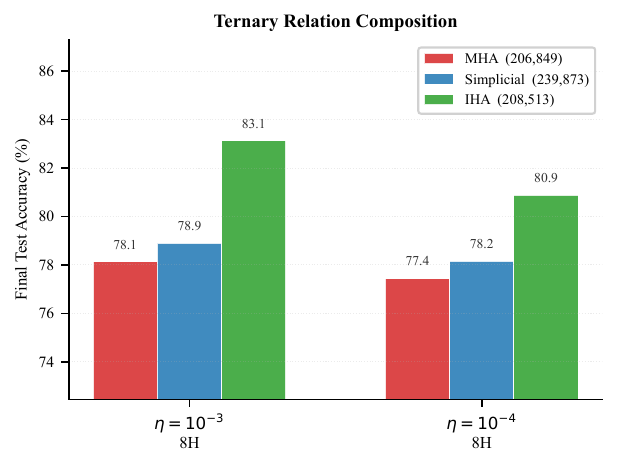}
    \caption*{(b) Ternary composition: final test accuracy across learning rates.}
    \label{fig:ternary-summary}
  \end{minipage}
  \caption{Final test accuracy summaries for binary and ternary relation composition. Bars compare MHA, IHA, and simplicial attention under $L=1$, $H=8$, for $\eta \in \{10^{-3}, 10^{-4}\}$.}
  \label{fig:final-test-summary}
\end{figure*}

\begin{figure*}[t]
  \centering
  \begin{minipage}[t]{0.49\textwidth}
    \centering
    \includegraphics[width=\linewidth]{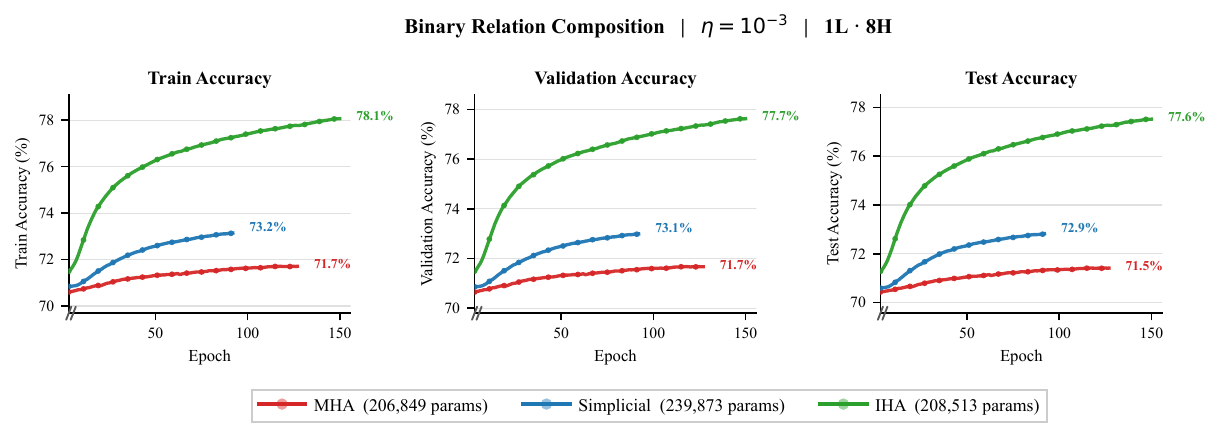}
    \caption*{(a) Binary composition, $\eta=10^{-3}$, $L=1$, $H=8$.}
    \label{fig:binary-curves-1e-3}
  \end{minipage}\hfill
  \begin{minipage}[t]{0.49\textwidth}
    \centering
    \includegraphics[width=\linewidth]{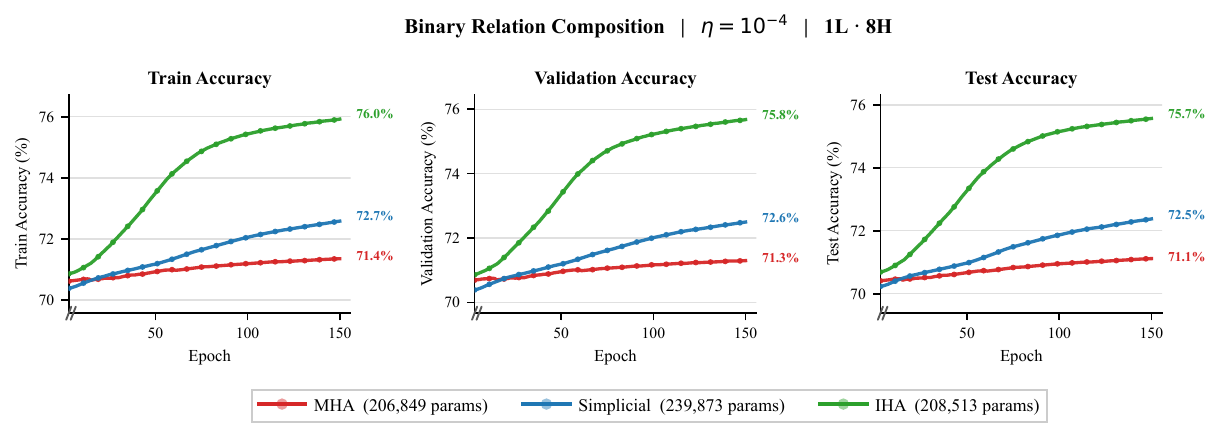}
    \caption*{(b) Binary composition, $\eta=10^{-4}$, $L=1$, $H=8$.}
    \label{fig:binary-curves-1e-4}
  \end{minipage}
  \caption{Learning curves for Binary Relation Composition. Each panel shows train, validation, and test accuracy versus epoch for MHA, IHA, and simplicial attention under a one-layer, eight-head transformer.}
  \label{fig:binary-learning-curves}
\end{figure*}

\begin{figure*}[t]
  \centering
  \begin{minipage}[t]{0.49\textwidth}
      \vspace{0pt}
    \centering
    \includegraphics[width=\linewidth]{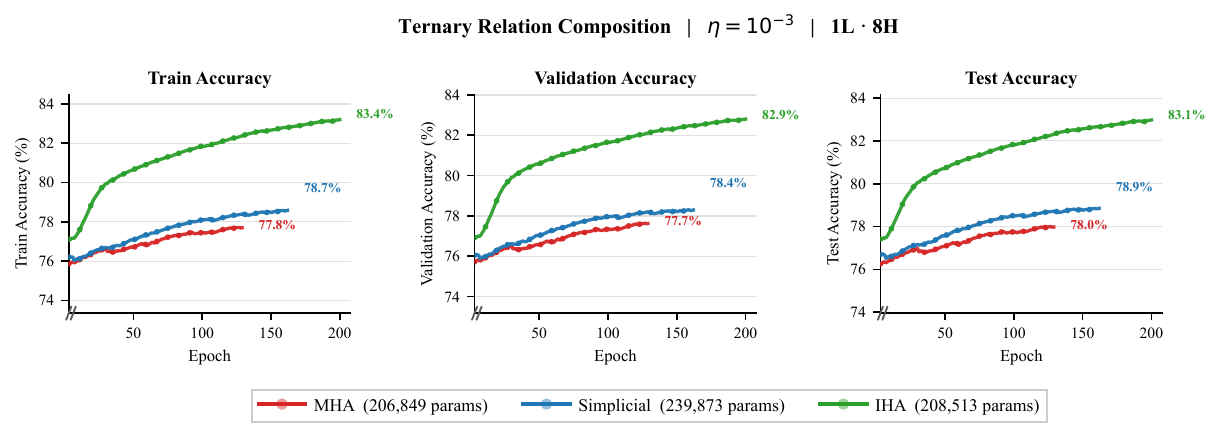}
    \caption*{(a) Ternary composition, $\eta=10^{-3}$, $L=1$, $H=8$.}
    \label{fig:ternary-curves-1e-3}
  \end{minipage}\hfill
  \begin{minipage}[t]{0.49\textwidth}
      \vspace{0pt}
    \centering
    \includegraphics[width=\linewidth]{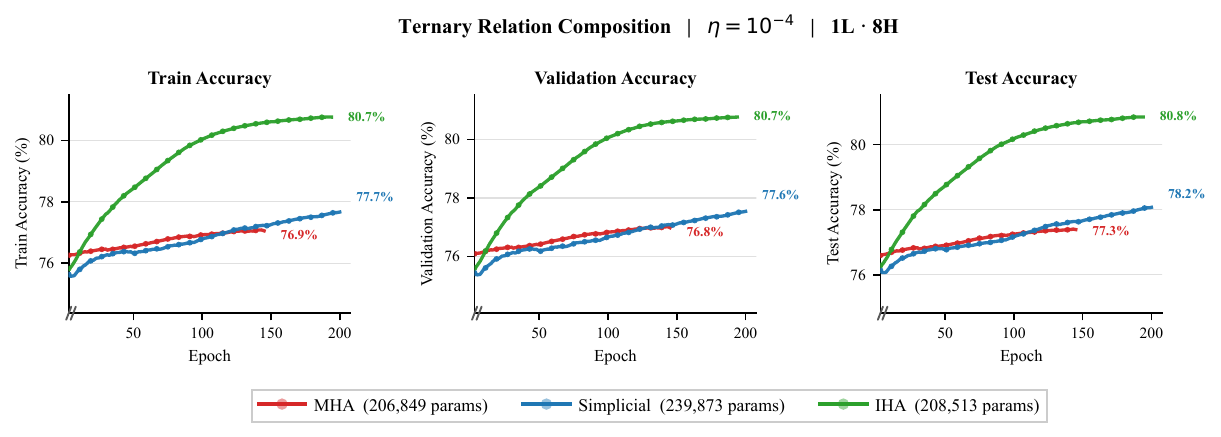}
    \caption*{(b) Ternary composition, $\eta=10^{-4}$, $L=1$, $H=8$.}
    \label{fig:ternary-curves-1e-4}
  \end{minipage}
  \caption{Learning curves for Ternary Relation Composition. Each panel shows train, validation, and test accuracy versus epoch for MHA, IHA, and simplicial attention under a one-layer, eight-head transformer.}
  \label{fig:ternary-learning-curves}
\end{figure*}

We evaluate multiple attention mechanisms including standard multi-head attention (MHA), interleaved head attention (IHA), and simplicial attention~\cite{roy2025fastsimplex}. All models use a single attention layer ($L=1$) with $8$ heads ($H=8$), and we sweep over two learning rates $\eta \in \{10^{-3}, 10^{-4}\}$. We use early stopping with a patience of 10 epochs. Note that for IHA, the number of pseudo, keys and values is $8$. In \autoref{fig:final-test-summary}, we report final test accuracy for each attention mechanism across the two learning rates. IHA consistently outperforms the other variants on both tasks and for both learning rates, achieving gains of up to $4.7\%$ on binary relation composition and $3.3\%$ on ternary relation composition relative to the strongest baseline, simplicial attention. For completeness, \autoref{fig:binary-learning-curves} and \autoref{fig:ternary-learning-curves} also show training, validation, and test accuracy as a function of epochs for each attention mechanism.

\end{document}